%% file: cvpr_arxiv.tex
\documentclass[10pt,twocolumn,letterpaper]{article}

\usepackage[pagenumbers]{cvpr} 

\usepackage{graphicx}
\usepackage{amsmath}
\usepackage{amssymb}
\usepackage{booktabs}
\usepackage{multirow}
\usepackage{indentfirst}

\usepackage[pagebackref,breaklinks,colorlinks]{hyperref}

\usepackage[capitalize]{cleveref}
\crefname{section}{Sec.}{Secs.}
\Crefname{section}{Section}{Sections}
\Crefname{table}{Table}{Tables}

\newcommand\blfootnote[1]{%
  \begingroup
  \renewcommand\thefootnote{}\footnote{#1}%
  \addtocounter{footnote}{-1}%
  \endgroup
}
 
\DeclareRobustCommand{\rchi}{{\mathpalette\irchi\relax}}
\newcommand{\irchi}[2]{\raisebox{\depth}{$#1\chi$}} 

\crefname{table}{Tab.}{Tabs.}

\newcommand\method{VaLID~} %

\def\cvprPaperID{6352} 
\def\confName{CVPR}
\def\confYear{2024}

\begin{document}

\title{VaLID: Variable-Length Input Diffusion for Novel View Synthesis }

\author{Shijie Li$^{1,2}$ $^*$ \quad
Farhad G. Zanjani$^{2}$\quad
Haitam Ben Yahia$^{2}$\\
Yuki M. Asano$^{2,3}$\quad
Juergen Gall$^{1}$\quad
Amirhossein Habibian$^{2}$ \vspace{-0.5em} \\\\
$^{1}$ University of Bonn
\qquad
$^{2}$ Qualcomm AI Research$^{\dagger}$
\qquad
$^{3}$ University of Amsterdam \\
\small
\texttt{\{lishijie,gall\}@iai.uni-bonn.de}\quad
\texttt{y.m.asano@uva.nl}\quad
\small
\texttt{\{fzanjani,hyahia,ahabibia\}@qti.qualcomm.com}
}


\input{macros}

\maketitle

\input{texts/0_abstract}
\blfootnote{$^*$Work completed during internship at Qualcomm Technologies, Inc.}
\blfootnote{$^{\dagger}$Qualcomm AI Research is initiative of Qualcomm Technologies,
Inc.}

\input{texts/1_introduction}

\input{texts/2_related_work}
\input{texts/3_method}
\input{texts/4_experiments}

\input{texts/5_conclusion}

\clearpage

{\small
\bibliographystyle{cvpr2023_stylekit/ieee_fullname}
\bibliography{bibliography}
}

\input{texts/supplementary_arxiv}

\end{document}

%% file: macros.tex

\newcommand{\head}[1]{{\smallskip\noindent\textbf{#1}}}
\newcommand{\alert}[1]{{\color{red}{#1}}}
\newcommand{\sm}{\scriptsize}
\newcommand{\eq}[1]{(\ref{eq:#1})}

\newcommand{\Th}[1]{\textsc{#1}}
\newcommand{\mr}[2]{\multirow{#1}{*}{#2}}
\newcommand{\mc}[2]{\multicolumn{#1}{c}{#2}}
\newcommand{\tb}[1]{\textbf{#1}}
\newcommand{\ul}[1]{\underline{#1}}
\newcommand{\ch}{\checkmark}

\newcommand{\red}[1]{{\color{red}{#1}}}
\newcommand{\blue}[1]{{\color{blue}{#1}}}
\newcommand{\green}[1]{\color{green}{#1}}
\newcommand{\gray}[1]{{\color{gray}{#1}}}

\newcommand{\citeme}[1]{\red{[XX]}}
\newcommand{\refme}[1]{\red{(XX)}}

\newcommand{\fig}[2][1]{\includegraphics[width=#1\linewidth]{fig/#2}}
\newcommand{\figh}[2][1]{\includegraphics[height=#1\linewidth]{fig/#2}}


\newcommand{\tran}{^\top}
\newcommand{\mtran}{^{-\top}}
\newcommand{\zcol}{\mathbf{0}}
\newcommand{\zrow}{\zcol\tran}

\newcommand{\ind}{\mathbbm{1}}
\newcommand{\expect}{\mathbb{E}}
\newcommand{\nat}{\mathbb{N}}
\newcommand{\zahl}{\mathbb{Z}}
\newcommand{\real}{\mathbb{R}}
\newcommand{\proj}{\mathbb{P}}
\newcommand{\prob}{\mathbf{Pr}}
\newcommand{\normal}{\mathcal{N}}

\newcommand{\mif}{\textrm{if}\ }
\newcommand{\other}{\textrm{otherwise}}
\newcommand{\minimize}{\textrm{minimize}\ }
\newcommand{\maximize}{\textrm{maximize}\ }
\newcommand{\st}{\textrm{subject\ to}\ }

\newcommand{\id}{\operatorname{id}}
\newcommand{\const}{\operatorname{const}}
\newcommand{\sgn}{\operatorname{sgn}}
\newcommand{\var}{\operatorname{Var}}
\newcommand{\mean}{\operatorname{mean}}
\newcommand{\trace}{\operatorname{tr}}
\newcommand{\diag}{\operatorname{diag}}
\newcommand{\vect}{\operatorname{vec}}
\newcommand{\cov}{\operatorname{cov}}
\newcommand{\sign}{\operatorname{sign}}
\newcommand{\prj}{\operatorname{proj}}

\newcommand{\softmax}{\operatorname{softmax}}
\newcommand{\clip}{\operatorname{clip}}

\newcommand{\defn}{\mathrel{:=}}
\newcommand{\peq}{\mathrel{+\!=}}
\newcommand{\meq}{\mathrel{-\!=}}

\newcommand{\floor}[1]{\left\lfloor{#1}\right\rfloor}
\newcommand{\ceil}[1]{\left\lceil{#1}\right\rceil}
\newcommand{\inner}[1]{\left\langle{#1}\right\rangle}
\newcommand{\norm}[1]{\left\|{#1}\right\|}
\newcommand{\abs}[1]{\left|{#1}\right|}
\newcommand{\frob}[1]{\norm{#1}_F}
\newcommand{\card}[1]{\left|{#1}\right|\xspace}
\newcommand{\divg}[2]{{#1\ ||\ #2}}
\newcommand{\diff}{\mathrm{d}}
\newcommand{\der}[3][]{\frac{d^{#1}#2}{d#3^{#1}}}
\newcommand{\pder}[3][]{\frac{\partial^{#1}{#2}}{\partial{#3^{#1}}}}
\newcommand{\ipder}[3][]{\partial^{#1}{#2}/\partial{#3^{#1}}}
\newcommand{\dder}[3]{\frac{\partial^2{#1}}{\partial{#2}\partial{#3}}}

\newcommand{\wb}[1]{\overline{#1}}
\newcommand{\wt}[1]{\widetilde{#1}}

\def\xssp{\hspace{1pt}}
\def\ssp{\hspace{3pt}}
\def\msp{\hspace{5pt}}
\def\lsp{\hspace{12pt}}

\newcommand{\cA}{\mathcal{A}}
\newcommand{\cB}{\mathcal{B}}
\newcommand{\cC}{\mathcal{C}}
\newcommand{\cD}{\mathcal{D}}
\newcommand{\cE}{\mathcal{E}}
\newcommand{\cF}{\mathcal{F}}
\newcommand{\cG}{\mathcal{G}}
\newcommand{\cH}{\mathcal{H}}
\newcommand{\cI}{\mathcal{I}}
\newcommand{\cJ}{\mathcal{J}}
\newcommand{\cK}{\mathcal{K}}
\newcommand{\cL}{\mathcal{L}}
\newcommand{\cM}{\mathcal{M}}
\newcommand{\cN}{\mathcal{N}}
\newcommand{\cO}{\mathcal{O}}
\newcommand{\cP}{\mathcal{P}}
\newcommand{\cQ}{\mathcal{Q}}
\newcommand{\cR}{\mathcal{R}}
\newcommand{\cS}{\mathcal{S}}
\newcommand{\cT}{\mathcal{T}}
\newcommand{\cU}{\mathcal{U}}
\newcommand{\cV}{\mathcal{V}}
\newcommand{\cW}{\mathcal{W}}
\newcommand{\cX}{\mathcal{X}}
\newcommand{\cY}{\mathcal{Y}}
\newcommand{\cZ}{\mathcal{Z}}

\newcommand{\vA}{\mathbf{A}}
\newcommand{\vB}{\mathbf{B}}
\newcommand{\vC}{\mathbf{C}}
\newcommand{\vD}{\mathbf{D}}
\newcommand{\vE}{\mathbf{E}}
\newcommand{\vF}{\mathbf{F}}
\newcommand{\vG}{\mathbf{G}}
\newcommand{\vH}{\mathbf{H}}
\newcommand{\vI}{\mathbf{I}}
\newcommand{\vJ}{\mathbf{J}}
\newcommand{\vK}{\mathbf{K}}
\newcommand{\vL}{\mathbf{L}}
\newcommand{\vM}{\mathbf{M}}
\newcommand{\vN}{\mathbf{N}}
\newcommand{\vO}{\mathbf{O}}
\newcommand{\vP}{\mathbf{P}}
\newcommand{\vQ}{\mathbf{Q}}
\newcommand{\vR}{\mathbf{R}}
\newcommand{\vS}{\mathbf{S}}
\newcommand{\vT}{\mathbf{T}}
\newcommand{\vU}{\mathbf{U}}
\newcommand{\vV}{\mathbf{V}}
\newcommand{\vW}{\mathbf{W}}
\newcommand{\vX}{\mathbf{X}}
\newcommand{\vY}{\mathbf{Y}}
\newcommand{\vZ}{\mathbf{Z}}

\newcommand{\va}{\mathbf{a}}
\newcommand{\vb}{\mathbf{b}}
\newcommand{\vc}{\mathbf{c}}
\newcommand{\vd}{\mathbf{d}}
\newcommand{\ve}{\mathbf{e}}
\newcommand{\vf}{\mathbf{f}}
\newcommand{\vg}{\mathbf{g}}
\newcommand{\vh}{\mathbf{h}}
\newcommand{\vi}{\mathbf{i}}
\newcommand{\vj}{\mathbf{j}}
\newcommand{\vk}{\mathbf{k}}
\newcommand{\vl}{\mathbf{l}}
\newcommand{\vm}{\mathbf{m}}
\newcommand{\vn}{\mathbf{n}}
\newcommand{\vo}{\mathbf{o}}
\newcommand{\vp}{\mathbf{p}}
\newcommand{\vq}{\mathbf{q}}
\newcommand{\vr}{\mathbf{r}}
\newcommand{\Vs}{\mathbf{s}}
\newcommand{\vt}{\mathbf{t}}
\newcommand{\vu}{\mathbf{u}}
\newcommand{\vv}{\mathbf{v}}
\newcommand{\vw}{\mathbf{w}}
\newcommand{\vx}{\mathbf{x}}
\newcommand{\vy}{\mathbf{y}}
\newcommand{\vz}{\mathbf{z}}

\newcommand{\vone}{\mathbf{1}}
\newcommand{\vzero}{\mathbf{0}}

\newcommand{\valpha}{{\boldsymbol{\alpha}}}
\newcommand{\vbeta}{{\boldsymbol{\beta}}}
\newcommand{\vgamma}{{\boldsymbol{\gamma}}}
\newcommand{\vdelta}{{\boldsymbol{\delta}}}
\newcommand{\vepsilon}{{\boldsymbol{\epsilon}}}
\newcommand{\vzeta}{{\boldsymbol{\zeta}}}
\newcommand{\veta}{{\boldsymbol{\eta}}}
\newcommand{\vtheta}{{\boldsymbol{\theta}}}
\newcommand{\viota}{{\boldsymbol{\iota}}}
\newcommand{\vkappa}{{\boldsymbol{\kappa}}}
\newcommand{\vlambda}{{\boldsymbol{\lambda}}}
\newcommand{\vmu}{{\boldsymbol{\mu}}}
\newcommand{\vnu}{{\boldsymbol{\nu}}}
\newcommand{\vxi}{{\boldsymbol{\xi}}}
\newcommand{\vomikron}{{\boldsymbol{\omikron}}}
\newcommand{\vpi}{{\boldsymbol{\pi}}}
\newcommand{\vrho}{{\boldsymbol{\rho}}}
\newcommand{\vsigma}{{\boldsymbol{\sigma}}}
\newcommand{\vtau}{{\boldsymbol{\tau}}}
\newcommand{\vupsilon}{{\boldsymbol{\upsilon}}}
\newcommand{\vphi}{{\boldsymbol{\phi}}}
\newcommand{\vchi}{{\boldsymbol{\chi}}}
\newcommand{\vpsi}{{\boldsymbol{\psi}}}
\newcommand{\vomega}{{\boldsymbol{\omega}}}

\newcommand{\rLambda}{\mathrm{\Lambda}}
\newcommand{\rSigma}{\mathrm{\Sigma}}

\newcommand{\vLambda}{\bm{\rLambda}}
\newcommand{\vSigma}{\bm{\rSigma}}


\makeatletter
\newcommand{\vast}[1]{\bBigg@{#1}}
\makeatother

\makeatletter
\newcommand*\bdot{\mathpalette\bdot@{.7}}
\newcommand*\bdot@[2]{\mathbin{\vcenter{\hbox{\scalebox{#2}{$\m@th#1\bullet$}}}}}
\makeatother

\makeatletter
\DeclareRobustCommand\onedot{\futurelet\@let@token\@onedot}
\def\@onedot{\ifx\@let@token.\else.\null\fi\xspace}

\def\eg{\emph{e.g}\onedot} \def\Eg{\emph{E.g}\onedot}
\def\ie{\emph{i.e}\onedot} \def\Ie{\emph{I.e}\onedot}
\def\cf{\emph{cf}\onedot} \def\Cf{\emph{Cf}\onedot}
\def\etc{\emph{etc}\onedot} \def\vs{\emph{vs}\onedot}
\def\wrt{w.r.t\onedot} \def\dof{d.o.f\onedot} \def\aka{a.k.a\onedot}
\def\etal{\emph{et al}\onedot}
\makeatother

%% file: texts/0_abstract.tex
\begin{abstract}
Novel View Synthesis (NVS), which tries to produce a realistic image at the target view given source view images and their corresponding poses, is a fundamental problem in 3D Vision.
As this task is heavily under-constrained, some recent work, like Zero123 \cite{liu2023zero}, tries to solve this problem with generative modeling, specifically using pre-trained diffusion models.
Although this strategy generalizes well to new scenes, compared to neural radiance field-based methods, it offers low levels of flexibility.
For example, it can only accept a single-view image as input, despite realistic applications often offering multiple input images.
This is because the source-view images and corresponding poses are processed separately and injected into the model at different stages.
Thus it is not trivial to generalize the model into multi-view source images, once they are available.
To solve this issue, we try to process each pose image pair separately and then fuse them as a unified visual representation which will be injected into the model to guide image synthesis at the target-views.
However, inconsistency and computation costs increase as the number of input source-view images increases.
To solve these issues, the Multi-view Cross Former module is proposed which maps variable-length input data to fix-size output data.
A two-stage training strategy is introduced to further improve the efficiency during training time.
Qualitative and quantitative evaluation over multiple datasets demonstrates the effectiveness of the proposed method against previous approaches. 
The code will be released according to the acceptance.


\end{abstract}

%% file: texts/1_introduction.tex
\section{Introduction}

Zero-shot novel view synthesis (NVS) has attracted more and more attention recently as the development of AR/VR applications \cite{kopf2020one} and content creation \cite{yen2022nerf,poole2022dreamfusion} rises.
NVS tries to synthesize high-quality and visually consistent views given source view images and relative poses.
Yet, because of the general under-constrainedness of reconstruction and due to occlusion, it is a highly challenging task that usually involves solving this large ambiguity. 

\begin{figure}[t]
\centering
\centering
\includegraphics[width=\linewidth]{{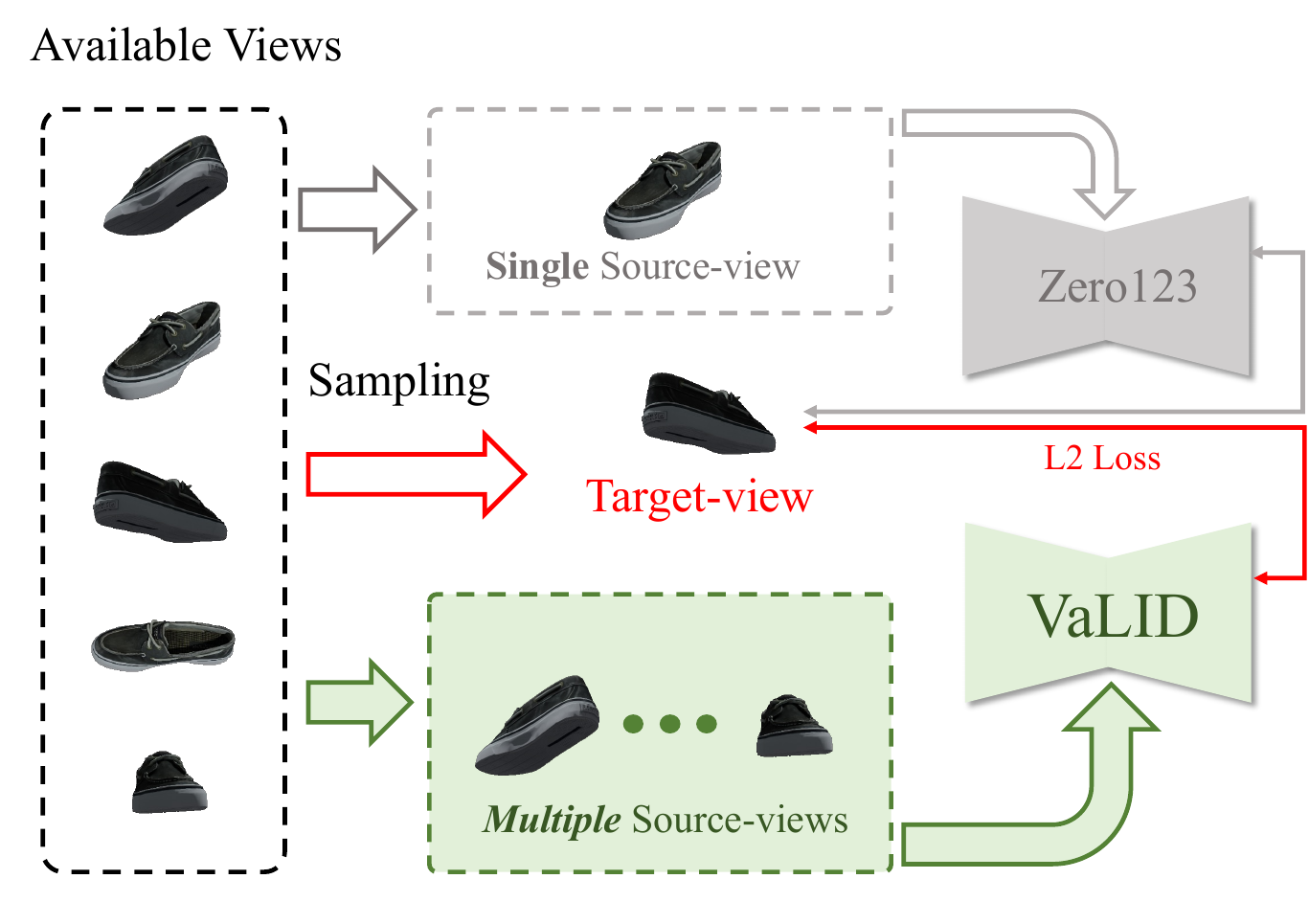}}
\vspace{-5mm}
\caption{Compared to the previous method (Zero123\cite{liu2023zero}), which can only receive a single image as input even when multiple images are available, the proposed architecture (\method) can accept variable input views in both training and inference. Thus it can learn a robust representation during training and can utilize more information in the inference time.}
\vspace{-3mm}
\label{fig:consistency}
\vspace{-5mm}
\end{figure}

Current solutions can be roughly partitioned into inference-time optimization-based methods and optimization-free methods.
The former~\cite{mildenhall2021nerf,niemeyer2022regnerf} can produce high-quality outcomes by conducting time-consuming iterative optimization for each object and thus cannot generalize to other objects directly.
As a comparison, the latter \cite{liu2023zero, melaskyriazi2023realfusion, liu2023syncdreamer, poole2022dreamfusion, weng2023consistent123, long2023wonder3d, shi2023MVDream, weng2023consistent123, yang2023consistnet, ye2023consistent} models the whole task by a neural network.
Although the generation is a single forward in the inference time which is highly efficient, the quality of generated images is not desirable.
Compared to the above methods, Zero123\cite{liu2023zero} shows a promising alternative solution.
By utilizing the powerful diffusion model, it can produce high-quality images with good generalization ability and efficiency.
However, it has been constrained to a single source-view image as input whereas, in the case of the availability of multi-view source images, it cannot be utilized in the process of NVS. The flexibility of a model to a variable number of source-view images is more desirable in real applications where more source images reduce ambiguity through multi-view fusion.
Figure~\ref{fig:overview}~(a) shows an overview of the Zero123\cite{liu2023zero} architecture. The model adopts an appearance-pose disentanglement conditioning strategy. The appearance information is injected at the input of U-Net whereas the corresponding relative pose information is injected in the attention modules in the U-Net. Due to separate conditioning of the diffusion process on appearance and pose embeddings, it is non-trivial to generalize the model to multi-view source images, once they are available.

To resolve this restriction and enable the model with the ability to receive variable size source-view images as input, an appearance-pose-entanglement conditioning strategy is proposed which is shown in Figure~\ref{fig:overview}~(b).
Based on this conditioning strategy, we propose a novel architecture named \textbf{Va}riable-\textbf{L}ength \textbf{I}nput \textbf{D}iffusion (VaLID).
Compared to prior works including Zero123\cite{liu2023zero}, \method can receive variable size source-view images as input to perform NVS, during both training and inference time. 
This flexibility allows to handle single-view and multi-view input at the same time and to produce even higher fidelity results when more
views are available.
The proposed \method first transfers the source-view image(s) into rich visual embeddings through a Vision Transformer (ViT) encoder, pre-trained with the Masked Auto-Encoder (MAE) task~\cite{he2022masked}. The ViT encoder produces multiple spatial output tokens. Similar to Zero123\cite{liu2023zero}, the camera pose(s) are transferred into the pose embedding(s) through an MLP network. The image tokens are concatenated with their corresponding camera pose features to form the input for conditioning the diffusion network.      

However, when multiple source-view images are available, there will be an inconsistency among tokens from different images. What is worse, the required computation resource will increase a lot due to the increased number of image tokens from multiple images.
To make tokens from different views work harmoniously while efficient, we propose a new  ``Multi-view Cross Former'' module.
In this module, all the tokens will be fused and transferred to a fixed number of tokens independent of the number of input source-view images.
Finally, these transferred tokens will be injected into the attention modules in the diffusion model to guide the target-view image synthesis.

Although the proposed architecture has the ability to receive multiple source-view images as input, this also requires more training resources.
To alleviate this issue, an efficient two-stage training strategy is introduced where the first stage mainly focuses on learning NVS from a single source-view image whereas the second stage mainly solves inconsistency when multiple source-view images are available by only finetuning relevant modules.
A token sampling strategy is applied at stage 2 to further constrain training cost. 

In summary, our contributions can be summarised as: 
\begin{itemize}
    \item We introduce a diffusion-based NVS model to address variable-size multi-view image fusion, both in training and inference times. The proposed appearance-pose-entanglement conditioning strategy can outperform previous methods quantitatively and qualitatively even when only a single source-view image is used.  
    \item We introduce Multi-view Cross Former to transfer the variable-size input tokens into a fixed-size representation by learning a set of learnable tokens. This improves the consistency and efficiency of conditioning to generate novel views while being agnostic to the number of input images.
    \item With the proposed two-stage training strategy, the training efficiency is improved. What's more, the performance of single-view image-conditioned NVS is also improved by involving multiple views in the training.
\end{itemize}

%% file: texts/2_related_work.tex
\section{Related work}
Novel View Synthesis is a fast-moving research area with many concurrent works. We broadly categorize the existing and concurrent works into single and multi-view models and highlight the key differences to our work.
\input{figs/arch}

\subsection{Single image novel-view synthesis}
Several NVS methods based on single-view input image have been presented on 
NeRF-based~\cite{mildenhall2021nerf} approaches to act on a single view~\cite{xu2022sinnerf, yu2021pixelnerf, xuneurallift360, melaskyriazi2023realfusion, szymanowicz23viewset_diffusion} to generate novel views. While these methods produce impressive results, NeRFs famously suffer from requiring test time optimization and the single-view variants typically do not work as well as their multi-view counterparts~\cite {yu2021pixelnerf}. To counter the lack of information from a single view, some include additional geometric priors as supervision, input or inductive bias, such as depth maps~\cite{xuneurallift360, xu2022sinnerf, tucker2020singleviewVS, han2022single, kant2023invs} point clouds ~\cite{mu20223d} or epipolar attention ~\cite{tseng2023poseguideddiffusion}. Our work, in contrast, does not use any geometric representation except for the relative pose. Although these geometric representations work well to aid in reconstructing the observed angle, it does not circumvent that some angles are simply occluded or unobserved. Therefore a large body of work aims at filling these gaps with generative models. Specifically, diffusion models ~\cite{ho2020denoising, sohl2015diffusion} have recently become the primary method to fill in gaps in the observed views as in RealFusion~\cite{melaskyriazi2023realfusion}, DreamFusion~\cite{poole2022dreamfusion} and Zero123~\cite{liu2023zero}. In particular, Zero-1-to-3~\cite{liu2023zero} has shown that this is possible with a single view and a relative camera pose. They use CLIP~\cite{radford2021learning} to extract appearance features, combined with a relative pose to generate novel views. Our work is closest to their setup, but we reduce high variance in unobserved regions by extending it to the multi-view setting for training and allowing for single or multi-view inference. This is the \emph{key difference} between single-view methods and ours: we investigate and quantify the effect of increasing the number of views, either during training or inference, without any geometric priors. Our work is orthogonal to many of the aforementioned methods and could be further improved as a combination.


\subsection{Multi-image novel view synthesis}
In this section, we look in particular at multi-view works in sparse view regimes. 
Similar to the single-view setup, there are several NeRF-based methods~\cite{jain2021putting, lin2023visionnerf, yariv2021volume, szymanowicz23viewset_diffusion, wang2021ibrnet, henzler2021UnsupervisedLO}, which try to mitigate the number of views needed for novel view synthesis.
Several papers focus on geometry-free NVS such as ENR~\cite{dupont2020equivariant} that lifts the extracted features of a pair input images into 3D space and minimizes the cross-rendering loss. The introduced 3D lifting is computationally expensive and degrades the quality of rendered images. LFNs~\cite{sitzmann2021lfns} samples a single point per ray cast and SRT~\cite{sajjadi2022srt} uses transformers to align multiple frames implicitly. Our work likewise is a geometry-free approach, but we additionally use generative modeling to synthesize novel views. EG3D~\cite{Chan2022} and 3DiM~\cite{watson2022novel}, similarly use generative modeling to align 3D features, using a GAN~\cite{goodfellow2014generative} or diffusion model respectively. 3DiM ~\cite{watson2022novel} is closest to our work with its diffusion model, but they only use multiple views during training time and generate poses in an auto-regressive manner at inference that suffers from accumulation of errors in generating and feeding in the additional views and being much slower than our work that consumes all the inputs in a single generation process.

MVDiffusion~\cite{tang2023mvdiffusion} uses text prompts, depth maps, pixel correspondences, multi-view, and diffusion models on scene reconstruction.
Magic3D~\cite{lin2023magic3d} trains a coarse NeRF with a diffusion model on top for high-resolution NVS based on a text prompt.
SyncDreamer~\cite{liu2023syncdreamer} does NVS from a single view, by lifting several noisy target images into a frustum and conditioning on this in a pre-trained Zero123. Several works try to enforce consistency across Zero-1-to-3 in different ways~\cite{ye2023consistent,weng2023consistent123,lin2023consistent123,yang2023consistnet}~\ie by projecting features to 3D~\cite{yang2023consistnet}. The closest to us is Consistent123~\cite{weng2023consistent123}, which also uses cross-attention among different views, but the main difference is that we look at single-view output rather than simultaneous multi-view output.
MVDREAM~\cite{shi2023MVDream} uses text prompts and diffusion models.
Wonder3D~\cite{long2023wonder3d} estimates normal maps and textures and combines them using optimization of a neural implicit signed distance
field (SDF) to amalgamate all 2D generated data.

Unlike the previously mentioned methods, the proposed \method does not require optimization during inference and can accommodate a varying number of source views, making it suitable for practical applications.


%% file: figs/arch.tex
\begin{figure*}
    \centering
    \resizebox{0.9\linewidth}{!}{
    \includegraphics{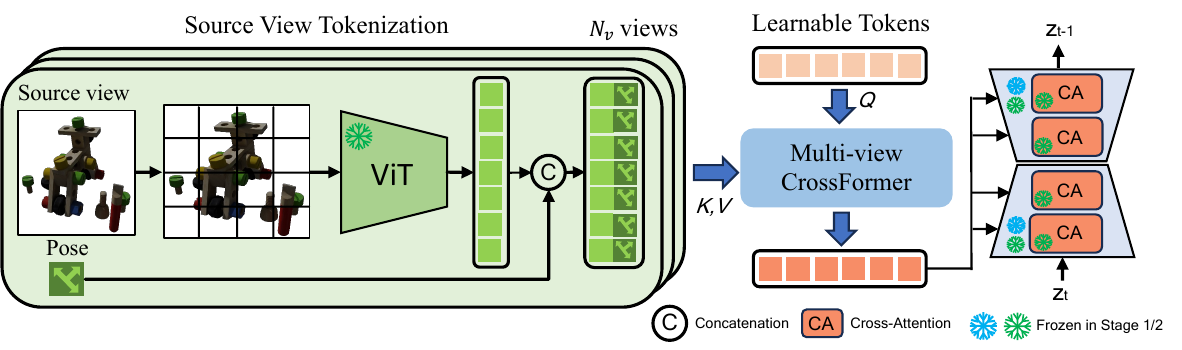}
    }
    \caption{Overview of the proposed architecture (\method).}
    \label{fig:arch}
    \vspace{-5mm}
\end{figure*}

%% file: texts/3_method.tex
\section{\method Novel View Synthesis}
We aim for generating a realistic image $\boldsymbol{y}$ at the target-view, given any number of source-view images $\boldsymbol{x} = \{ x_i\}_{i=1}^{N_v}$, and corresponding camera poses relative to the target-view $\boldsymbol{\pi} = \{ \pi_i\}_{i=1}^{N_v}$, where $N_v\geq1$ is a variable number of input source-view images.
\method involves two steps: \emph{i)} Extracting features from the source-view images as described in Sec. \ref{sec:feature_extraction}. \emph{ii) } Aggregating features from source-view images to generate the target-view as discussed in Sec. \ref{sec:multi_view_fusion}. A high-level overview of our pipeline is illustrated in Figure \ref{fig:arch}.

\subsection{Source view tokenization}
\label{sec:feature_extraction}

\input{figs/overview}
\paragraph{Limitations of existing conditionings}
The prior work Zero123\cite{liu2023zero} is made of two conditioning mechanisms as illustrated in Fig~\ref{fig:overview} (a): \emph{i)} \emph{U-Net conditioning}, where a source-view image $x_i$ will go through a frozen Auto-Encoder. The output latent map $f^{AE}_i$ is then concatenated with a noisy latent feature map $z_t$ from the previous diffusion timestep $t+1$ as input to the U-Net. \emph{ii)} \emph{Attention conditioning}, where the source-view image $x_i$ will go through a frozen CLIP image encoder. The output CLIP embedding $f^\mathrm{CLIP}_i$ is concatenated with the relative pose $\pi_i$ and then fed into the attention modules in the U-Net. Zero123\cite{liu2023zero} can thus be formulated as:
\begin{equation}
z_{t-1}=\boldsymbol{\epsilon}_\theta\left(f^{AE}_i \oplus z_t, {f^\mathrm{CLIP}_i} \oplus\pi_i, t\right), 
\label{eq:z123}
\end{equation}
where $\boldsymbol{\epsilon}_\theta$ denotes the U-Net,  $t$ the timestep in the diffusion process and symbol $\oplus$ denotes the concatenation operator.
%
We found that the Attention conditioning mainly ignores the CLIP features and only propagates pose information as shown in Figure~\ref{fig:motivation}. More examples are available in the supplemental materials.  We hypothesize that the CLIP encoding, a high-level single token image embedding, is too coarse to retain image details. Thus Eq.~(\ref{eq:z123}) reverts to $z_{t-1} \approx \boldsymbol{\epsilon}_\theta\left(f^{AE}_i \oplus z_t, \pi_i, t\right)$, where we see that the extracted features of input source-view image $f^{AE}_i$ and relative pose $\pi_i$ are injected into the U-Net at different stages.
This works well when only a single source-view image is available as in Zero123\cite{liu2023zero}.
However, when multiple source-view images are available, this disentangled appearance-pose conditioning strategy would require the model to align the poses with the source-view features at a later stage in the network. 
\\
\textbf{Learning joint appearance and pose conditioning.} We propose to entangle appearance and pose features together. Since only U-Net conditioning encodes appearance information, one straightforward solution is that we feed both source-view images and corresponding poses to the U-Net input. The problem here is that the number of input views would be inherently tied to the first convolutional layer in the U-Net, making it inflexible. Furthermore, when multiple source-view images are available, feeding them directly into the U-Net will produce inconsistent results, which are shown in Figure~\ref{fig:inconsistency}.
Thus we utilize the flexibility of the attention mechanism and opt for feeding both source-view images and corresponding poses to the attention modules in U-Net.
In this way, we need to replace the CLIP image encoder by another architecture to better encode source-view images.
We achieve this with the $\mathbf{SVT}$ Module which can extract fine spatial information from source-view images.
Our final architecture can be formulated as:

\begin{equation}
z_{t-1}=\boldsymbol{\epsilon}_\theta\left(z_{t}, \mathbf{SVT}(x_1, \pi_1, ..., x_{N_{v}} , \pi_{N_{v}}), t\right).
\end{equation}

Then the network is trained by optimizing a simplified variational lower-bound

\begin{equation}
    \mathcal{L} = \mathbb{E}[||\boldsymbol{\epsilon}_t - z_{t-1}||^2].
\end{equation}

where $\boldsymbol{\epsilon}_t \sim \mathcal{N}(0, 1)$.

%

\input{figs/motivation}

\input{figs/consistency}  

\paragraph{Source View Tokenization Module}
In the Source View Tokenization ($\mathbf{SVT}$) Module, each input source-view pose image $(x_i, \pi_i)$ is processed separately and converted into pose-image tokens encoding both appearance and geometry information.
We extract features for each input source-view image $x_i$ with a ViT encoder ($\Phi$), pre-trained with the Masked Auto-Encoder (MAE) task~\cite{he2022masked}.
The input image will be  converted into small image patches $\boldsymbol{\rchi}= \{x_j\}_{j=1}^{N_{p}}$ where $N_{p}$ is the number of image patches. Positional embeddings are added to the images as in \cite{dosovitskiy2020image} for spatial awareness.
 These image patches are then fed into the ViT encoder and converted into image tokens:

\begin{equation}
    \mathbf{M} =  \Phi(\boldsymbol{\rchi}), \quad
    \mathbf{M} = \{m_j\}_{j=1}^{N_p}
\end{equation}
At this stage, each image token $m_j$ only contains appearance and 2D spatial information whereas camera pose information is missing.
To inject such information, the 3D poses are appended to the end of each token:
\begin{equation}
    \mathbf{\mathbf{\tilde{M}}} = \{ \tilde{m}_j \}_{j=1}^{N_p}, \text{where } \tilde{m}_j = m_j \oplus \pi^s .
\end{equation}
where $\pi^s$ is the corresponding pose to each image token $m_j$
For now, each pose-image token $\tilde{m}_j$ is aware of both appearance information and spatial information (2D spatial and 3D pose).
Compared to the CLIP image encoder, the Source View Tokenization can extract rich spatial information and thus better describe image details. The proposed architecture is shown in Figure \ref{fig:arch}.
Later, the extracted tokens from multiple source views can be fused together in our Multi-view Token Cross Former module described in the next Section.

\subsection{Multi View Fusion}
\label{sec:multi_view_fusion}
\subsubsection{Multi-view Cross Former}
In the Source View Tokenization Module, we extract pose-image tokens $\mathbf{\tilde{M}}_i$ for each input source-view image.
As mentioned earlier, inconsistent results are produced when providing multi-views directly to the U-Net (Figure~\ref{fig:inconsistency}).
Alternatively feeding all source view image tokens will increase the computation proportionally to the number of images, and
causes out-of-memory issues. Inspired by text-to-image diffusion models~\cite{rombach2022high}, where a fixed number of tokens have been shown to perform favorably, we introduce a fixed number (equal to 64) of learnable seed tokens $S_0$, initialized as $S_0 \sim \mathcal{N}(0,1)$ at the beginning of training. These are used as queries in the Multi-view Cross Former block, visualized in Figure \ref{fig:mta}. Since these initial tokens $S_0$ are shared across training examples, they learn to be query token biases, which aim to extract relevant information from pose-image tokens $\mathbf{\tilde{M}}_i$.
The Multi-view Cross Former block, in particular, is first computed with an attention operation followed by a residual:
\begin{equation}
    \mathbf{S}'_{l} = \mathbf{Attn}_{l}(Q, K, V) + \mathbf{S}_{l-1}.
\end{equation}
Then the output tokens will pass through a feed-forward layer $\mathbf{FFW}_{l}$ which consists of layer normalization, a linear layer, GeLU activation, another linear layer sequentially, followed by a final residual connection:

\begin{equation}
    \mathbf{S}_l = \mathbf{FFW}_{l}(\mathbf{S}_l') + \mathbf{S}_l'.
\end{equation}

The query, key, and value in the attention block are calculated with $Q = \mathbf{S}_{l-1}$ and $\textbf{}K = V = \mathbf{\tilde{M}}_1 \oplus ... \oplus \mathbf{\tilde{M}}_{N_v}$.
The output tokens $\mathbf{S}_l$ will be used as new target-view seeds for the following layer and the whole block will be repeated $L$ times. 
Finally, the target-view seeds $\mathbf{S}_L$  will be fed into attention modules at each timestep of the diffusion process to produce realistic images at the target view.
With this design, the final target-view seeds $\mathbf{S}_L$ learn to warp the source view tokens into the target-view tokens that are being used for conditioning the diffusion process.
This also enables our model to always feed a fixed number of query tokens into the U-Net attention module
which improves efficiency a lot.

\input{figs/mta}





\subsubsection{Efficient Training and Inference}
\label{sec:training}
As feeding multiple source-view images during training is still expensive, we split the whole training procedure into two steps.
In the first stage, the proposed architecture is trained to produce target-view images given a single source-view image.
After this stage, our model already has the ability to produce realistic target-view images.
As multi-view information aggregation only takes effect in the Multi-view Cross Former, we freeze all modules apart from the Multi-view Cross Former in the second stage of the training.
With this strategy in the second stage of training, all the training resources will be used to make multi-view information work harmoniously to produce consistent outcomes. 
Additionally, we propose a sampling strategy during the second stage training where we feed a variable number of views to the model and randomly sample pose-image tokens from them.
By reducing the number of pose-image tokens fed into the Multi-view Cross Former, the training cost is further reduced.

\textbf{Stage 1: Single-view Optimization} We build the proposed model on top of a pretrained stable-diffusion model. Thus we can assume that the U-Net has already the ability to produce realistic images. With this assumption, we only optimize the attention modules in the U-Net.
Thus in this stage, the Source View Tokenization Module, Multi-view Cross Former, and attention modules in the U-Net are optimized with a single source-view image as input.

\begin{figure*}[t]
    \centering
    \resizebox{\linewidth}{!}{
    \includegraphics{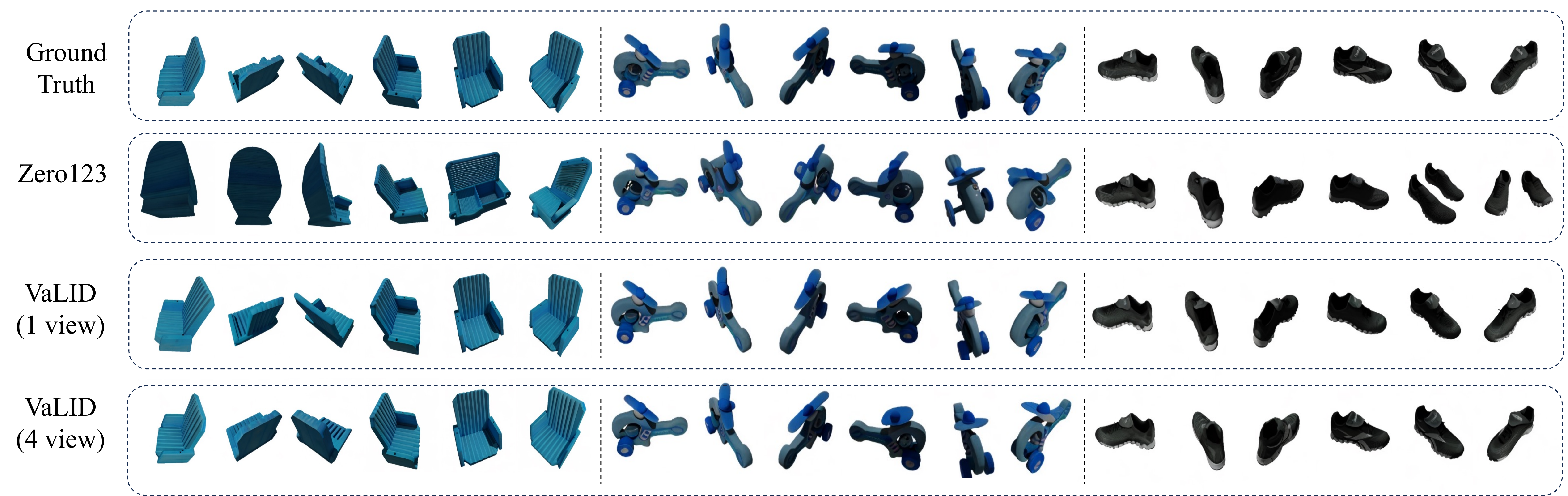}
    }
    \vspace{-5mm}
    \caption{Qualitative results on GSO dataset. We can observe our method can produce high-quality outcomes at the target view compared to the previous state-of-the-art methods. Furthermore, the generated images from our method also maintain consistency across different views. This is demonstrated by generated images along a predefined camera trajectory. Especially when multiple source-view images are available, the quality of generated images improves a lot.
    More examples can be found in the supplemental materials.
    }
    \label{fig:qvis_gso}
\end{figure*}

\begin{table*}[h]
    \centering
    \setlength{\tabcolsep}{3mm}
    \resizebox{0.9\linewidth}{!}{
        \begin{tabular}{l|cccc|cccc}
        \hline & DietNeRF~\cite{jain2021putting} & IV~\cite{sdiv} & SJC-I~\cite{wang2023score}  & Zero123 \cite{liu2023zero} &  \multicolumn{4}{c}{\method} \\
        \hline
        View num & 1 & 1 & 1 & 1 & 1 & 2 & 3 & 4 \\
        \hline PSNR $\uparrow$ & $ 8.933$ & 5.914 & 6.573 & 19.000 &  20.034 & 20.405 & 21.053 & 21.305 \\
        SSIM $\uparrow$ & 0.645 & 0.540 & 0.552 &  0.865 & 0.881 & 0.884 & 0.891 & 0.895 \\
        LPIPS $\downarrow$ & $ 0.412 $ & 0.545 & 0.484 & 0.115 & 0.091 & 0.085 & 0.073 & 0.069 \\
        \hline
        \end{tabular}
        }
    \caption{NVS results on Google Scanned Objects dataset. The proposed method outperforms the previous state-of-the-art method by a  significant margin on all metrics. Furthermore, as the number of input source-view images increased, the performance gradually improved.}
    \label{tab:exp_gso}
    \vspace{-3mm}
\end{table*}

\begin{table}[h]
    \centering
    \setlength{\tabcolsep}{3mm}
    \resizebox{\linewidth}{!}{
        \begin{tabular}{l|cccc|cccc}
        \hline & DietNeRF~\cite{jain2021putting} & IV~\cite{sdiv} & SJC-I~\cite{wang2023score} & Zero123 \cite{liu2023zero}  & \method \\
        \hline
        Input views & 1 & 1 & 1 & 1 & 1\\
        \hline PSNR $\uparrow$ & 7.130 & 6.561 & 7.953  & 8.893 & 9.024\\
        SSIM $\uparrow$ & 0.406 & 0.442 & 0.456  & 0.515 & 0.519\\
        LPIPS $\downarrow$ & 0.507 & 0.564 & 0.545  & 0.432 & 0.432\\
        \hline
        \end{tabular}
        }
        
    \caption{NVS results on RTMV dataset. Compared to the Google Scanned Objects dataset, this dataset contains more challenging scenes. We can observe our method still can outperform other methods.}
    \label{tab:exp_rtmv}
    \vspace{-7mm}
\end{table}

\textbf{Stage 2: Variable-view Optimization} After stage 1 training, we only finetune the Multi-view Cross Former where multi-view information fusion takes place, highlighted in Figure~\ref{fig:arch}. In each training iteration, variable input views are fed into the model while only partial pose-image tokens are fed into the Multi-view Cross Former for optimization. With this design, we not only improve the training efficiency due to reducing the number of used pose-image tokens but also improve the robustness of the proposed architecture when only partial information is available.
At inference time, we can choose to feed all pose-image tokens to the Multi-view Cross Former to provide more information to the model. We can alternatively sample fewer tokens for efficiency purposes 
though we find this also hurts performance (shown in Sec.~\ref{sec:ablation}).

%% file: figs/overview.tex

\begin{figure*}[h]
\centering
\subfloat[Conditioning in Zero123\cite{liu2023zero}]{
\begin{minipage}[t]{0.4\linewidth}
\centering
\includegraphics[width=\linewidth]{{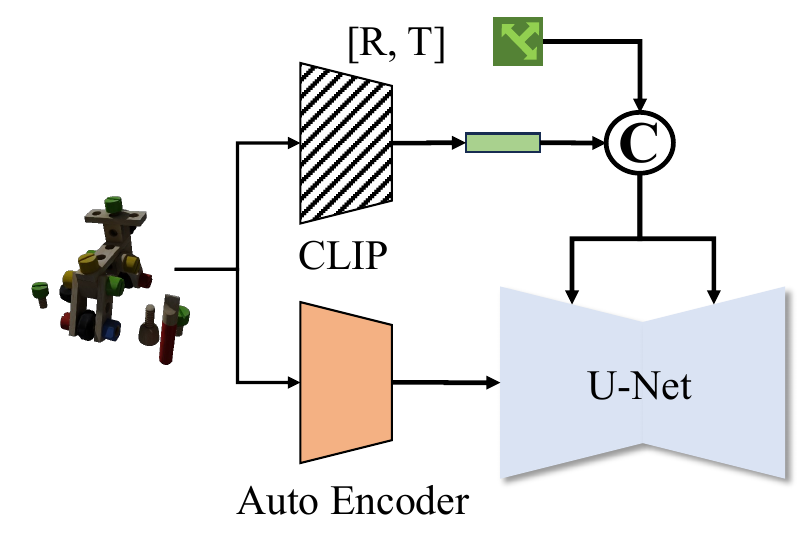}}
\end{minipage}
}
\vspace{5mm}
\subfloat[Conditioning in \method]{
\begin{minipage}[t]{0.5\linewidth}
\centering
\includegraphics[width=\linewidth]{{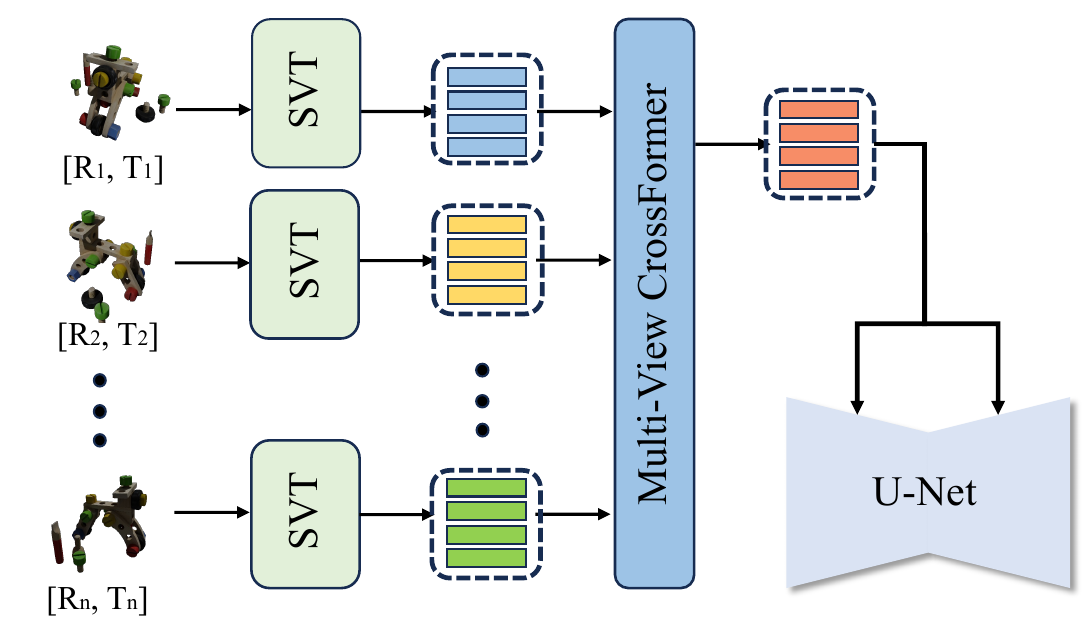}}
\end{minipage}
}
\vspace{-5mm}
\caption{
View conditioning in prior work (a) vs. in \method (b). By learning a joint appearance pose representation, \method seamlessly handles variable length input, which is not possible in a disentangled representation of appearance and pose as in Zero123\cite{liu2023zero}.}
\vspace{-5mm}
\label{fig:overview}
\end{figure*}   

%% file: figs/motivation.tex
\begin{figure}[t!]
\centering
\hspace{-5mm}
\subfloat[Target]{
\begin{minipage}[t]{0.25\linewidth}
\centering
\includegraphics[width=\linewidth]{{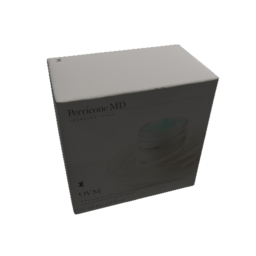}}\\
\vspace{-3mm}
\includegraphics[width=\linewidth]{{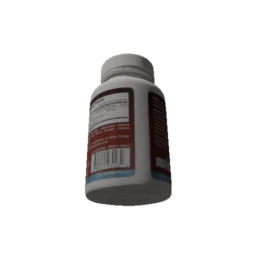}}\\
\vspace{-3mm}
\includegraphics[width=\linewidth]{{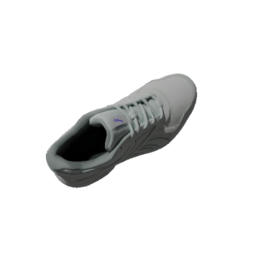}}
\end{minipage}
}
\hspace{-5mm}
\subfloat[Zero123\cite{liu2023zero}]{
\begin{minipage}[t]{0.25\linewidth}
\centering
\includegraphics[width=\linewidth]{{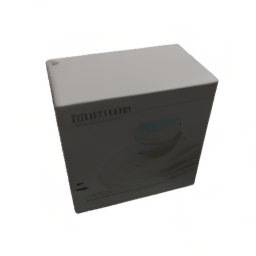}}\\
\vspace{-3mm}
\includegraphics[width=\linewidth]{{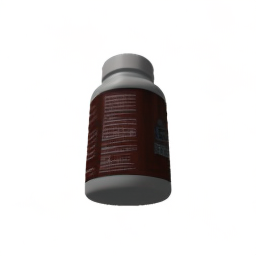}}\\
\vspace{-3mm}
\includegraphics[width=\linewidth]{{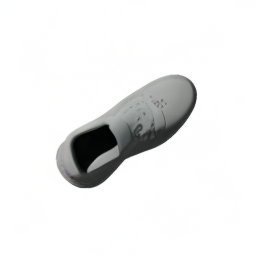}}
\end{minipage}
}
\hspace{-5mm}
\subfloat[w/o CLIP]{
\begin{minipage}[t]{0.25\linewidth}
\centering
\includegraphics[width=\linewidth]{{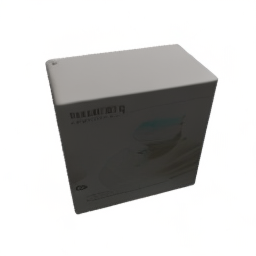}}\\
\vspace{-4mm}
\includegraphics[width=\linewidth]{{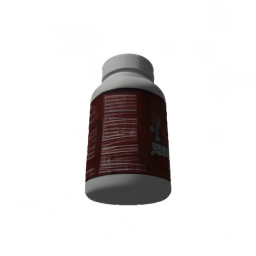}}\\
\vspace{-4mm}
\includegraphics[width=\linewidth]{{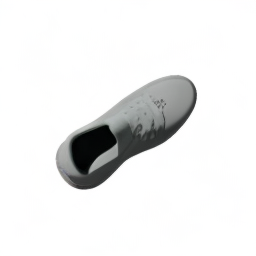}}
\end{minipage}
}
\hspace{-5mm}
\subfloat[w/o UC]{
\begin{minipage}[t]{0.25\linewidth}
\centering
\includegraphics[width=\linewidth]{{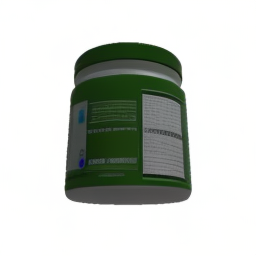}}\\
\vspace{-3mm}
\includegraphics[width=\linewidth]{{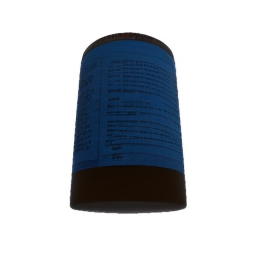}}\\
\vspace{-3mm}
\includegraphics[width=\linewidth]{{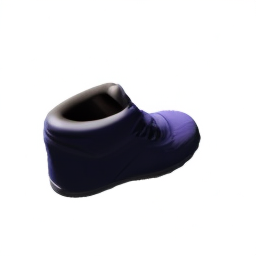}}
\end{minipage}
}

\caption{
Conditioning limitations in Zero123~\cite{liu2023zero}: Conditioning is dominated by the U-Net input (UC) whereas CLIP embedding is ignored (CLIP).
}
\label{fig:motivation}
\vspace{-5mm}
\end{figure}

%% file: figs/consistency.tex
\begin{figure}[t]
\centering
\hspace{-5mm}
\subfloat[1 view]{
\begin{minipage}[t]{0.25\linewidth}
\centering
\includegraphics[width=\linewidth]{{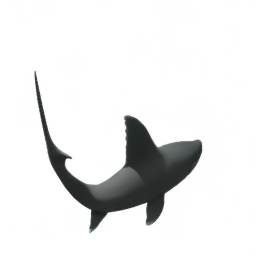}}\\
\vspace{-1mm}
\includegraphics[width=\linewidth]{{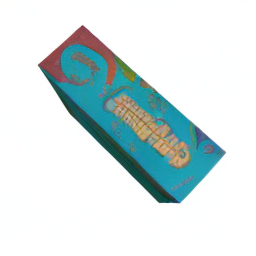}}\\
\vspace{-3mm}
\includegraphics[width=\linewidth]{{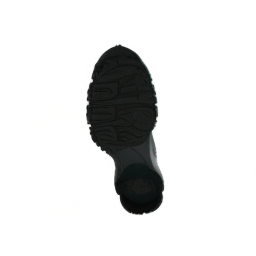}}
\end{minipage}
}
\hspace{-5mm}
\subfloat[2 views]{
\begin{minipage}[t]{0.25\linewidth}
\centering
\includegraphics[width=\linewidth]{{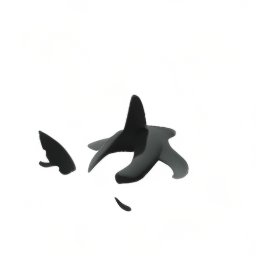}}\\
\vspace{-1mm}
\includegraphics[width=\linewidth]{{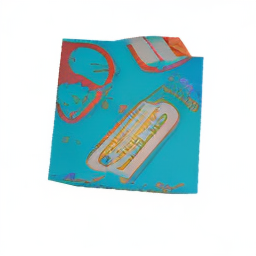}}\\
\vspace{-3mm}
\includegraphics[width=\linewidth]{{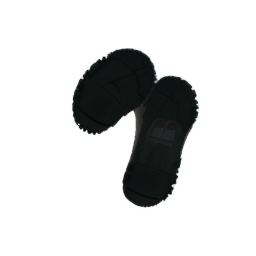}}
\end{minipage}
}
\hspace{-5mm}
\subfloat[3 views]{
\begin{minipage}[t]{0.25\linewidth}
\centering
\includegraphics[width=\linewidth]{{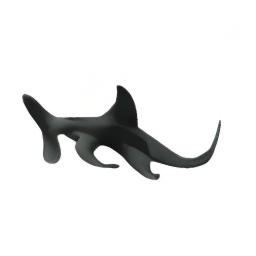}}\\
\vspace{-1mm}
\includegraphics[width=\linewidth]{{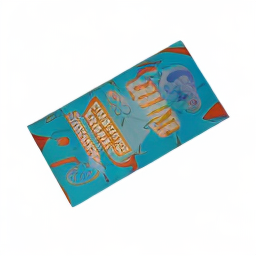}}\\
\vspace{-3mm}
\includegraphics[width=\linewidth]{{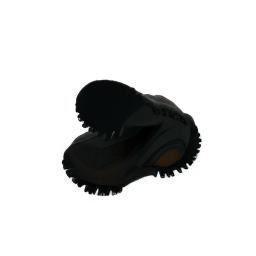}} \\
\end{minipage}
}
\hspace{-5mm}
\subfloat[4 views]{
\begin{minipage}[t]{0.25\linewidth}
\centering
\includegraphics[width=\linewidth]{{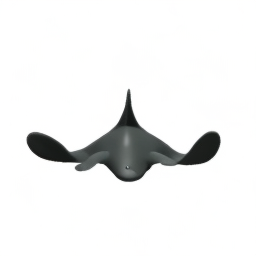}}\\
\vspace{-1mm}
\includegraphics[width=\linewidth]{{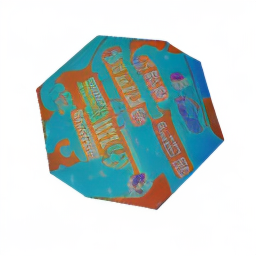}}\\
\vspace{-3mm}
\includegraphics[width=\linewidth]{{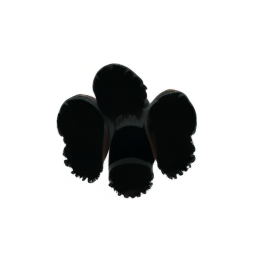}}
\end{minipage}
}
\vspace{-3mm}
\caption{We can observe without the stage 2  training, there will exist heavily inconsistency in the outcomes when we feed multiple source-view images to the proposed architecture.}
\label{fig:inconsistency}
\vspace{-5mm}
\end{figure} 

%% file: figs/mta.tex
\begin{figure}[t]
    \centering
    \resizebox{0.9\linewidth}{!}{
    \includegraphics{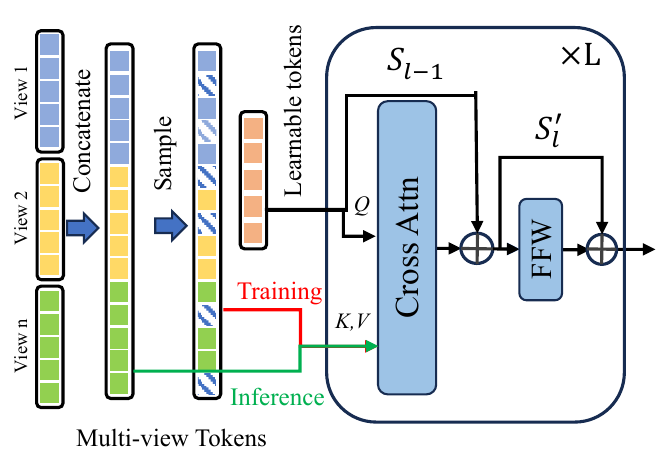}
    }
    \caption{Multi-view Cross Former. During \textcolor{red}{\textbf{training}} time (stage 2), we randomly sample image tokens from all available source-view images. With this strategy, we can make the Multi-view Cross Former receives multi-view information while restrict training cost. In the \textcolor{green}{\textbf{inference}} time, we either feed all the tokens to include all information or sample as during training.
    }
    \vspace{-5mm}
    \label{fig:mta}
\end{figure}

%% file: texts/4_experiments.tex
\section{Experiments}
\subsection{Experimental Setting}
\textbf{Dataset} We follow Zero123's\cite{liu2023zero} experimental setting and fine-tune our diffusion model on the Objaverse dataset \cite{deitke2023objaverse} which contains more than 800K high-quality 3D object CAD models.
For a fair comparison, we use the processed rendering data provided by Zero123\cite{liu2023zero} which are 12 random views for each object.
For evaluation, we mainly focus on the performance of out-of-distribution data.
Thus we conducted the evaluation on the Google Scanned Objects dataset \cite{downs2022google} (GSO), which contains high-quality scanned household items, and RTMV \cite{tremblay2022rtmv} dataset, which provides more complex scenes each with around 20 random objects.
In both training and evaluation, all the images are resized to $256\times256$ resolution with a white background.
To be compliant with the visual transformer architecture in the Source View Tokenization Module, we conduct the center crop thus the final input size will be 224$\times$224.\\
\textbf{Evaluation Setting} On the Google Scanned Objects dataset, 24 views are randomly sampled in 3D as the target view.
To demonstrate the performance of our method with variable-length input source-view images, multiple source-view images are rendered following some predefined rules.
A reference view (view 0) is randomly picked up first with polar angle 60$^{\circ}$ for each object.
Then 3 other views (view 1,2,3) will be clockwisely selected by only varying azimuth (azimuth angle interval as 90$^{\circ}$).
In this setting, the available input information is controlled increasing gradually proportional to the number of input source-view images.
Because the RTMV dataset already provides rendered images, we conducted the evaluation following the Zero123 \cite{liu2023zero} experimental setting for fair comparison.
To quantitatively evaluate the quality of generated images, we use Peak signal-to-noise ratio (PSNR), Structural Similarity Index (SSIM) \cite{wang2004image}, and Learned Perceptual Image Patch Similarity (LPIPS) \cite{zhang2018unreasonable} to measure similarity between rendered images and ground truth images.
\begin{figure*}[t]
\centering
\subfloat[PSNR]{
\begin{minipage}[t]{0.25\linewidth}
\centering
\includegraphics[width=\linewidth]{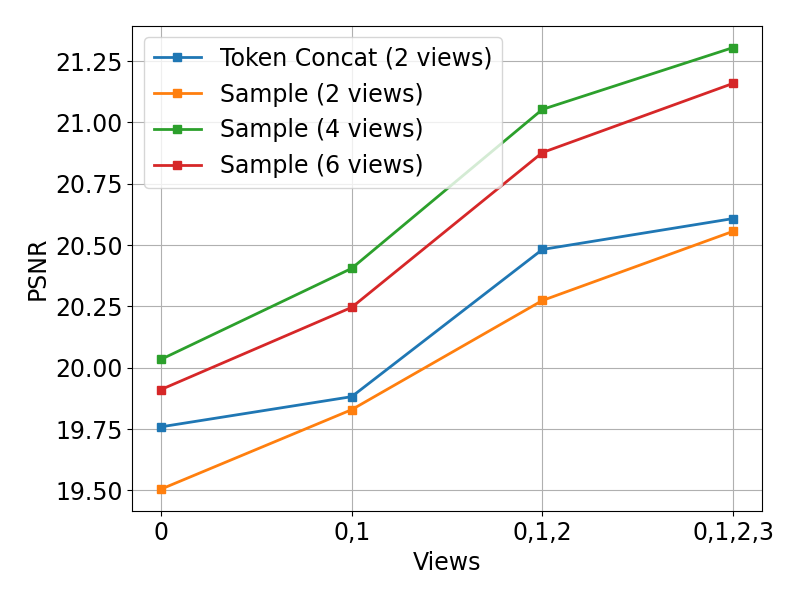}  
\end{minipage}
}
\subfloat[SSIM]{
\begin{minipage}[t]{0.25\linewidth}
\centering
\includegraphics[width=\linewidth]{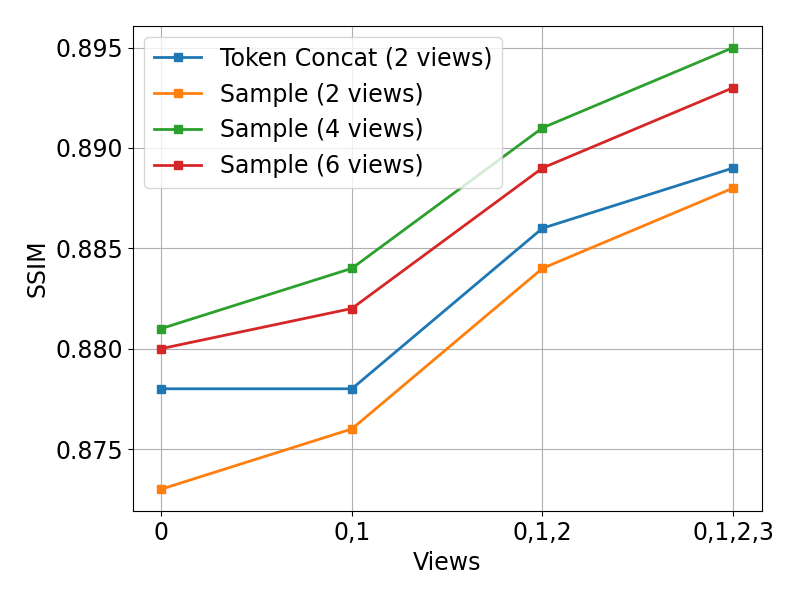}  
\end{minipage}
}
\subfloat[LPIPS]{
\begin{minipage}[t]{0.25\linewidth}
\centering
\includegraphics[width=\linewidth]{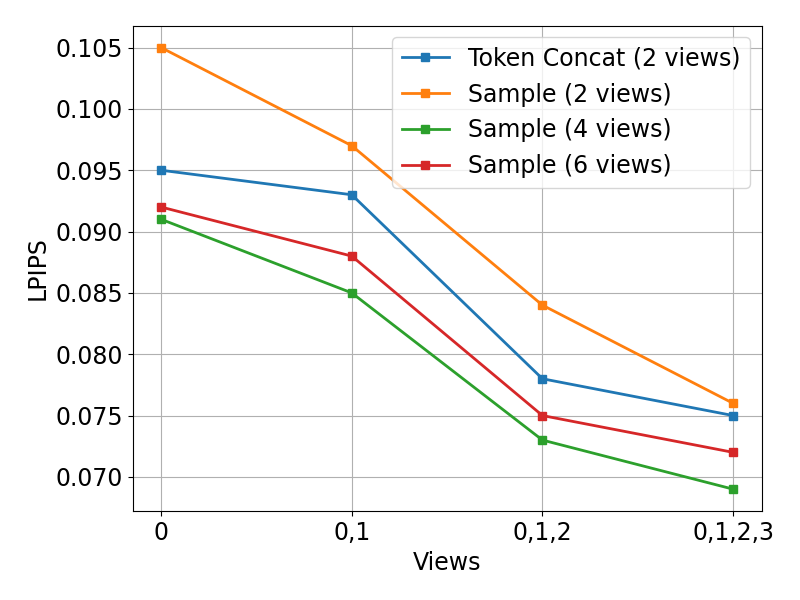}  
\end{minipage}
} 
\subfloat[Efficiency]{
\begin{minipage}[t]{0.25\linewidth}
\centering
\includegraphics[width=\linewidth]{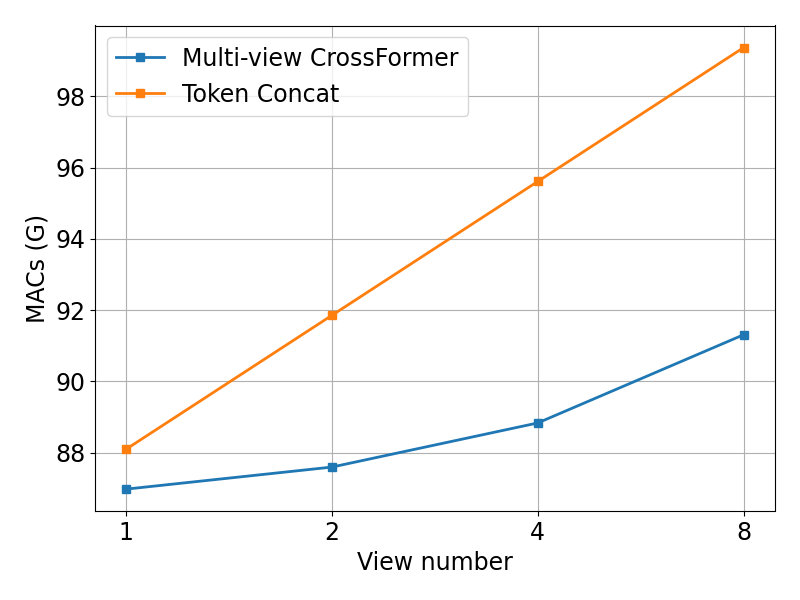}  
\end{minipage}
} 
\vspace{-3mm}
\caption{Token Sampling setting in the stage 2 training (a)-(c). The efficiency analysis of the proposed architecture is shown in (d).}
\vspace{-3mm}
\label{fig:sampling_training}
\end{figure*}
\begin{figure*}[t]
\centering
\subfloat[Multi-view Comparison]{
\begin{minipage}[t]{0.25\linewidth}
\centering
\includegraphics[width=\linewidth]{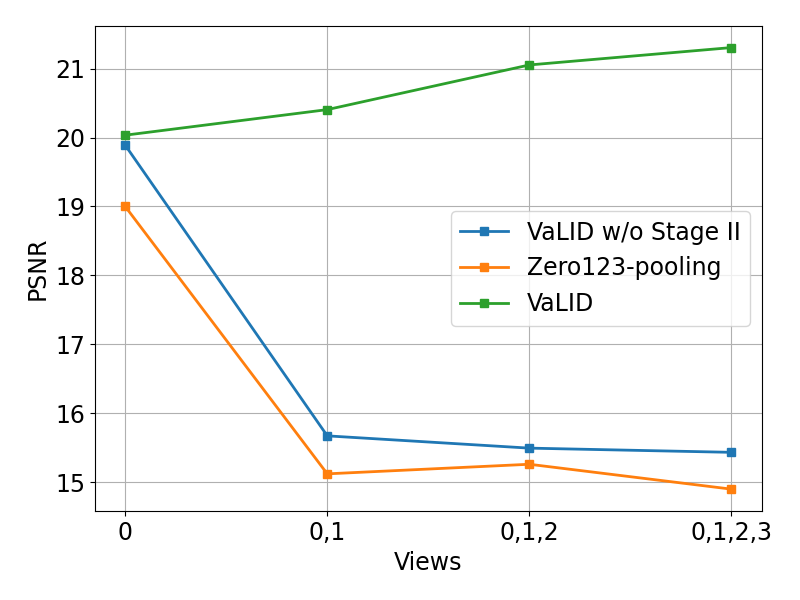}  
\end{minipage}
} 
\subfloat[PSNR]{
\begin{minipage}[t]{0.25\linewidth}
\centering
\includegraphics[width=\linewidth]{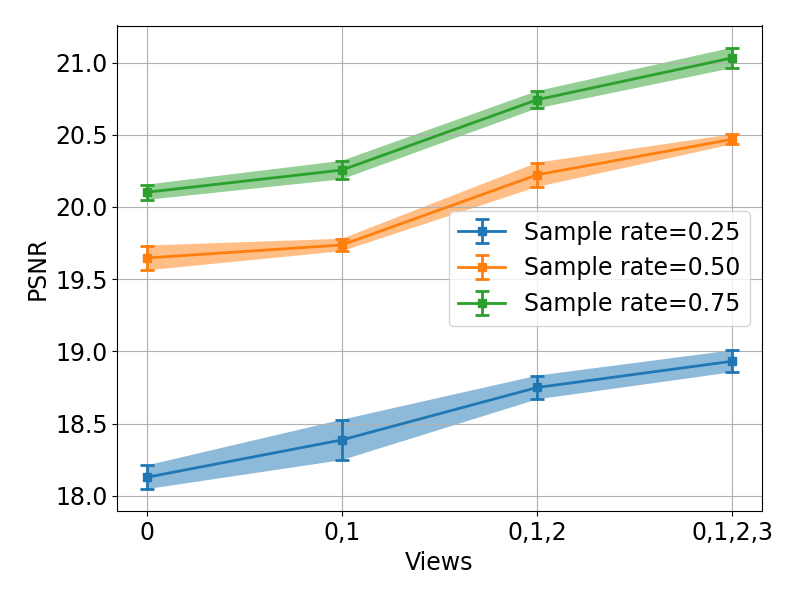}  
\end{minipage}
}
\subfloat[SSIM]{
\begin{minipage}[t]{0.25\linewidth}
\centering
\includegraphics[width=\linewidth]{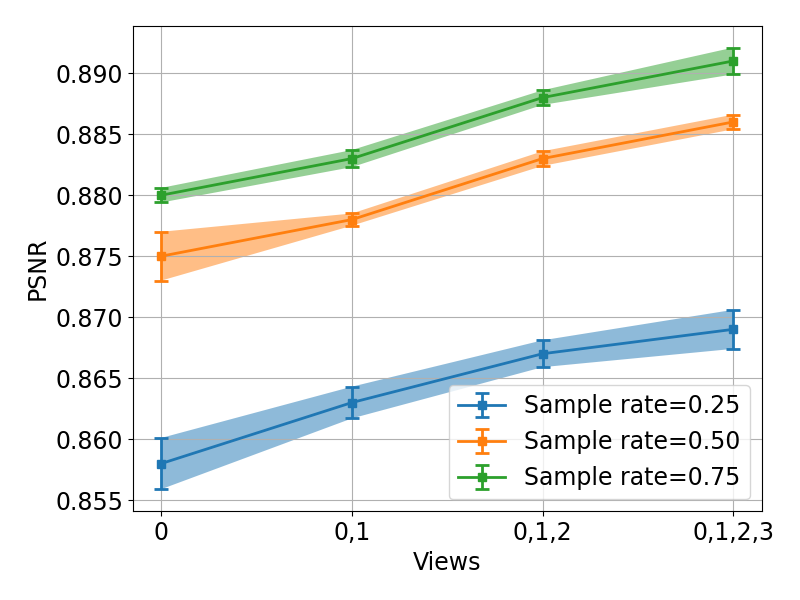}  
\end{minipage}
}
\subfloat[LPIPS]{
\begin{minipage}[t]{0.25\linewidth}
\centering
\includegraphics[width=\linewidth]{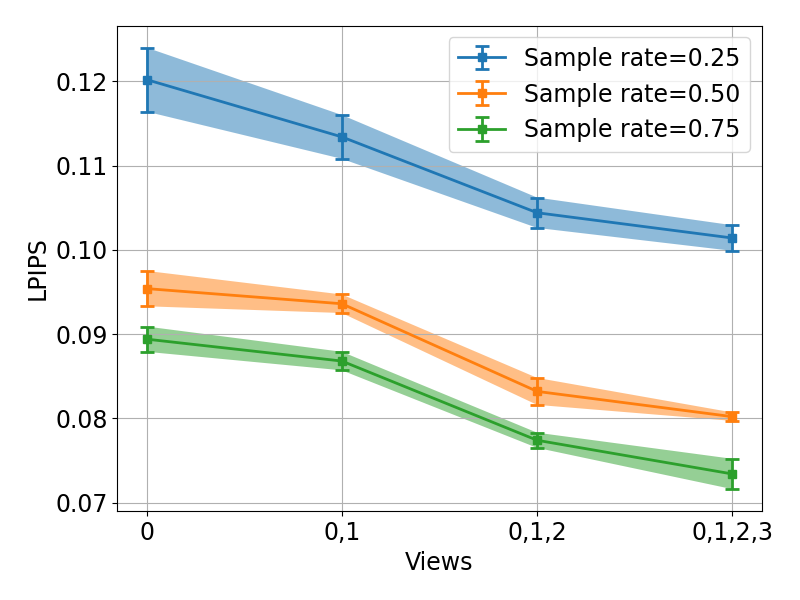}  
\end{minipage}
} 
\vspace{-3mm}
\caption{We show the necessity of stage 2 training in (a). The effect of token sampling at Inference time is shown in (b)-(d). \textcolor{green}{Green line} denotes sampling 75$\%$ tokens before Multi-view Cross Former. \textcolor{red}{Red line} and \textcolor{blue}{Blue line} represent sample ratio equals to 50\% and 25\% respectively. We can observe as the available information increase, the uncertainty decrease which are measured by 5 runs. 
}
\vspace{-7mm}
\label{fig:sampling_inference}
\end{figure*}
\\
\textbf{Baselines}
We mainly compare our method to the previous state-of-the-art method, Zero123\cite{liu2023zero} which can produce realistic images at the target view iteratively with a diffusion model.
Besides, an image-conditioned diffusion model Image Variations (IV)\cite{sdiv} is also selected as the baseline method which can produce semantic consistent images based on input images.
This is achieved by finetuning a Stable Diffusion model to be conditioned on images instead of text prompts.
Although Neural Radiance Field \cite{mildenhall2021nerf} (NeRF) is widely adopted for novel view synthesis, it only works well when a large number of images are available.
Hence, to further evaluate the effectiveness of our proposed method, we compare it to DietNeRF~\cite{jain2021putting}, a technique that regularizes NeRF using a CLIP image-to-image consistency loss. DietNeRF has demonstrated its capability to produce high-quality results even when only a limited number of images are available.
Additionally, zero-shot 3D content creation, which optimizes a NeRF with a pretrained diffusion model becomes popular. We use SJC\cite{wang2023score} which combines an image-conditioned diffusion model and NeRF as another baseline method to represent these methods.

\subsection{Comparison to the state-of-the-art methods}

We first compare our method to the baseline methods quantitatively on Google Scanned Objects and RTMV datasets.
The results are shown in Tab.~\ref{tab:exp_gso} and Tab.~\ref{tab:exp_rtmv}.
For the GSO dataset, the proposed method can achieve a 5\% improvement on PSNR when only a single source-view image is used. The improvement on SSIM and LPIPS are also obvious.
By increasing the input source-view images, the performances on all metrics increase
gradually. This demonstrates the capability of the proposed method to utilize information from multiple source-view images when available.

Furthermore, the proposed architecture has the flexibility to receive variable-length inputs and thus is more suitable for realistic applications or when the model receives input images progressively.
As for the RTMV dataset, we follow the Zero123\cite{liu2023zero} experiment setting because this dataset already provides rendered images.
For each scene, the first view is used as the source view and the following views are used as the target view.
Compared to the GSO dataset, this dataset contains more objects for each scene thus more challenging.
Even in this situation, our method can still outperform other baseline methods.

Finally, we show some qualitative results in Figure~\ref{fig:qvis_gso}.
Apart from the quality of each individually generated image, the consistency among different target views is also important which is usually missing in previous work.
To demonstrate the consistency of the generated images among different target views, we select a predefined camera trajectory to demonstrate the qualitative results.
From these qualitative results, we can observe the proposed method not only can produce more realistic images compared to previous methods but also can maintain consistency among different target views even no explicit geometry constraints are applied.
Especially when multiple source-view images are available.

\subsection{Ablation Study}
\label{sec:ablation}
In the ablation study, we show the influence of sampling strategy in both inference and training times.
First, we show the effect of the sampling strategy, involved in stage 2 of training in Figure~\ref{fig:sampling_training} (a)-(c).
It can be observed that in stage 2 of training, tokens sampling from two source views actually achieve quite close performance, compared to feeding all the tokens to Multi-view Cross Former.
Moreover, the experiments show that providing more source-view images improves the quality of the generated images. However, increasing up to six views doesn't show improvement in comparison to four views. This may be because a fixed number of learnable tokens is applied in Multi-view Cross Former. When the number of input tokens becomes too large, it will face a bottleneck to collect all useful information.

The computational burden of preserving (without sampling) all pose-image tokens for different numbers of source-view images is shown in Figure~\ref{fig:sampling_training}~(d). It can be observed that the introduced token sampling strategy shows a significant reduction in the required Multiply-Accumulate Operations (MACs) for U-Net and Multi-view Cross Former, especially when the number of source views increased. Thanks to Multi-view Cross Former, ignoring a considerable number of pose-image tokens via sampling does not show a performance loss while it obtains high efficiency in computation. 



The importance of the stage 2 training is shown in Figure \ref{fig:sampling_inference}~(a).
Without the stage 2 training, when multiple source-view images are available, the performance of our method decreases a lot. 
This is mainly because of inconsistency in multi-view tokens which is shown in Figure \ref{fig:inconsistency} previously.
As for Zero123\cite{liu2023zero}, we apply a pooling operation to enable it with the ability to receive multiple source-view images. Unfortunately, the performance also decreases significantly when multiple source-view images are fed.
Another interesting finding is that stage 2 training can also improve the performance of single source-view image-conditioned NVS.
This is because receiving multi-view information during training enables the model to learn a more comprehensive representation even if only a single source-view image is available in inference time.
This ablation shows the impact of stage 2 training where the Multi-view Cross Former adapts to perform multi-view token fusion when its input contains pose-image tokens belonging to distinct views.

In inference time, we try to validate the robustness of the proposed method.
This is achieved by feeding partial tokens into Multi-view Cross Former in the inference time similar to above.
The results are shown in Figure \ref{fig:sampling_inference} (b)-(d). The number is reported on 5 runs.
It can be observed that when the number of available source-view images increases or a higher number of tokens is used, the performance improves gradually.
Furthermore, the uncertainty is reduced gradually as the available information increases.
These results also show the robustness of the proposed method as even though limited information is available, the performance is still at a high level.

%% file: texts/5_conclusion.tex
\section{Conclusion}

This paper presents a novel diffusion-based framework for novel view image synthesis, named as Variable-Length Input Diffusion model (VaLID).
Compared to previous diffusion-based methods, which can only receive a single image as input, VaLID can accept variable number of views as input in both training and inference time.
This flexibility makes the proposed model more suitable for realistic applications where multiple views but with a variable number are usually available or progressively presents to model.
The multi-view fusion is achieved by an appearance-pose entanglement conditioning strategy.
To handle the information inconsistency among different views, a Multi-view Cross Former module has been introduced which is highly efficient, both in terms of reducing the number of tokens, which conditioned the diffusion process, to have a fix-length set and in terms of training strategy. 
The proposed method outperforms previous methods both quantitatively and qualitative which demonstrates its effectiveness.


%% file: texts/supplementary_arxiv.tex
\maketitlesupplementary

\input{figs/supp_zero123_cond}

\section{Conditioning Strategy}

As mentioned in the main paper, Zero123\cite{liu2023zero} is made of two conditioning mechanisms; (1) \emph{U-Net conditioning}, where a source-view image $x_i$ will go through a frozen Auto-Encoder and the output latent map $f^{AE}_i$ is then concatenated with a noisy latent feature map $z_t$ from the previous diffusion timestep $t+1$ as input to the U-Net. (2) \emph{Attention conditioning}, where the source-view image $x_i$ will go through a frozen CLIP image encoder. The output CLIP embedding $f^\mathrm{CLIP}_i$ is concatenated with the relative pose $\pi_i$ and then fed into the attention modules in the U-Net.

Figure~\ref{fig:cond} demonstrates the impacts of these two conditioning strategies on the generated images. 
To drop out the CLIP embedding, we mask them out with a 0-tensor of the same size.
It can be observed that in the case of following the default setting, where both U-Net conditioning and Attention conditioning exist, the model can produce plausible outcomes.
As a comparison, removing U-Net conditioning (w/o concat) produces poor outcomes, e.g. the objects in the generated images are usually in the wrong pose.
Moreover, the appearance of objects in generated images looks vastly different from the corresponding objects in the source-view images.
We hypothesize this is because the CLIP Image encoder can only output a single token for each input image which is a high-level semantic summary.
Thus, it is usually not enough to maintain image details.
By removing attention conditioning, we find the outcomes are almost the same, compared to the default setting.
This demonstrates U-Net conditioning dominates outcomes of Zero123 whereas CLIP embedding is almost ignored. 

\section{Qualitative Results}

Figure~\ref{fig:qvis_1} shows more qualitative results. It can be observed that compared to Zero123, our method can produce high-quality images.
In some examples, Zero123 produces objects in the wrong pose, the wrong shapes, or multiple objects whereas only a single object exists in the source-view images.
This may be because there exists high uncertainty when only a single source-view image is available.
Unfortunately, Zero123 cannot handle this uncertainty well, especially when the difference between source-view and target-view is large.
Intuitively, the uncertainty usually decreases as the available information (source-view images) increases.
By receiving variable-length input views, the proposed method \emph{VaLID} can utilize multi-view image information thus producing high-quality images.
We can observe a clear improvement when the number of input views increases. 
Even if only a single source-view image is available, it can still outperform Zero123 qualitatively. 

To further demonstrate the utilization and fusion of multi-view input images by the proposed \emph{VaLID} method, Figure \ref{fig:stage} shows the generated images with or without stage 2 training.
As it can be observed in these examples, after stage 1 training, the model has the ability to produce plausible outcomes (see column (b)).
Although our model in stage 1 has the flexibility to receive variable-length input views, it has not been trained to fuse multi-view inputs. So, at this stage, the inference on multi-view inputs shows inconsistency in data fusion to generate reasonable output (see columns (c-e)).
In other words, since the training inputs in stage 1 of training always contain a single view image, the multi-view Cross Former block has not been adapted to perform the multi-view fusion task.
To empower the model to perform the multi-view fusion, in stage 2 of training where only Cross Former parameters are tuned, the variable number of views are introduced as inputs.
With this efficient strategy, Multi-view Cross Former learns how to combine provided information from multiple images to generate a consistent output image.
As the number of input source-view images increases, the quality of produced images will gradually increase (see columns (f-i)).

Finally, we show some qualitative results in the attached video to show our method can produce more consistent images compared to previous methods.
These videos consist of generated images at a predefined camera trajectory.
We can observe with the single source-view image as input, that our method already can produce more consistent outcomes. When more input views are available, the consistency is improved further.












\input{figs/supp_qvis_1}

\input{figs/supp_stage}

%% file: figs/supp_zero123_cond.tex
\begin{figure*}[!t]
\centering
\hspace{-5mm}
\subfloat[Source View]{
\begin{minipage}[t]{0.2\linewidth}
\centering
\includegraphics[width=\linewidth]{{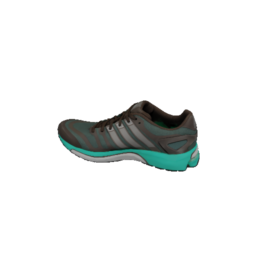}}\\
\includegraphics[width=\linewidth]{{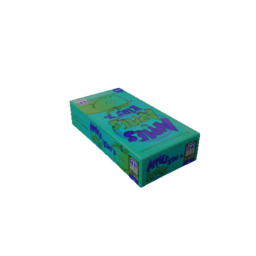}}\\

\includegraphics[width=\linewidth]{{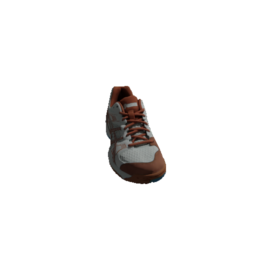}}\\

\includegraphics[width=\linewidth]{{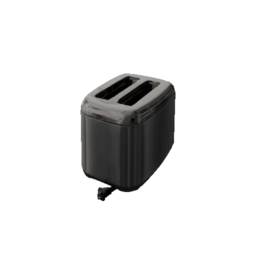}}\\

\includegraphics[width=\linewidth]{{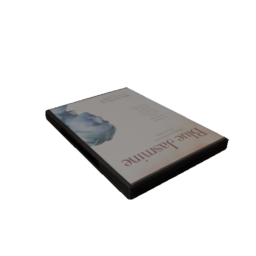}}\\
\end{minipage}
}
\hspace{-5mm}
\subfloat[Target View]{
\begin{minipage}[t]{0.2\linewidth}
\centering
\includegraphics[width=\linewidth]{{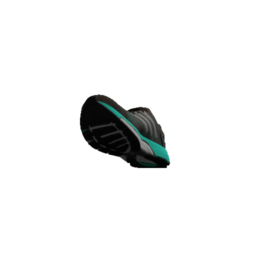}}\\
\includegraphics[width=\linewidth]{{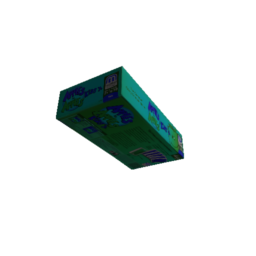}}\\

\includegraphics[width=\linewidth]{{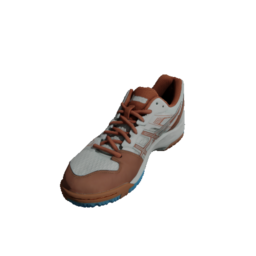}}\\

\includegraphics[width=\linewidth]{{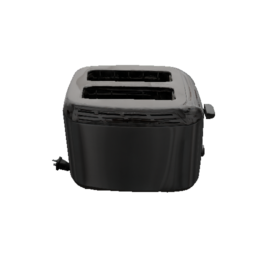}}\\

\includegraphics[width=\linewidth]{{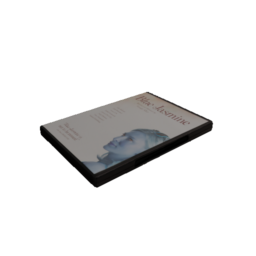}}\\
\end{minipage}
}
\hspace{-5mm}
\subfloat[Zero123\cite{liu2023zero}]{
\begin{minipage}[t]{0.2\linewidth}
\centering
\includegraphics[width=\linewidth]{{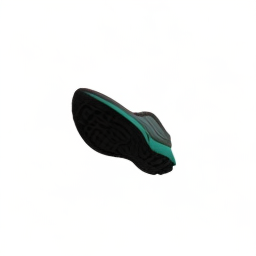}}\\

\includegraphics[width=\linewidth]{{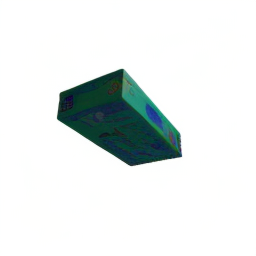}}\\

\includegraphics[width=\linewidth]{{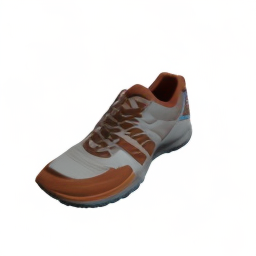}}\\

\includegraphics[width=\linewidth]{{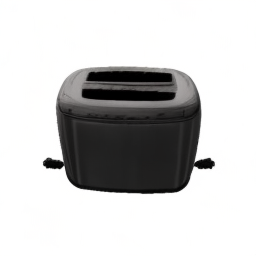}}\\

\includegraphics[width=\linewidth]{{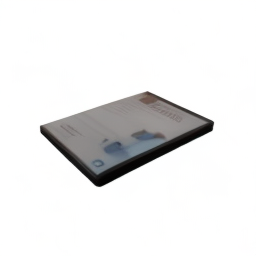}}\\
\end{minipage}
}
\hspace{-5mm}
\subfloat[Zero123 w/o concat]{
\begin{minipage}[t]{0.2\linewidth}
\centering
\includegraphics[width=\linewidth]{{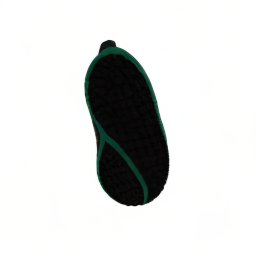}}\\
\includegraphics[width=\linewidth]{{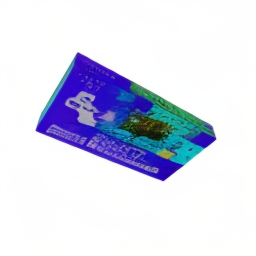}}\\

\includegraphics[width=\linewidth]{{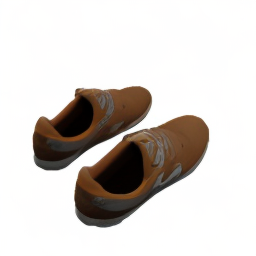}}\\

\includegraphics[width=\linewidth]{{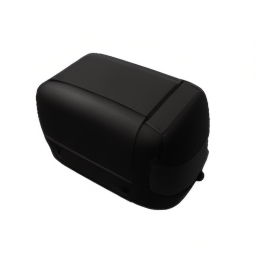}}\\

\includegraphics[width=\linewidth]{{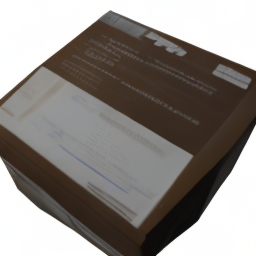}}\\
\end{minipage}
}
\subfloat[Zero123 w/o CLIP]{
\begin{minipage}[t]{0.2\linewidth}
\centering
\includegraphics[width=\linewidth]{{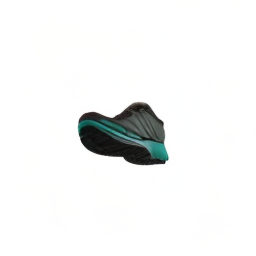}}\\
\includegraphics[width=\linewidth]{{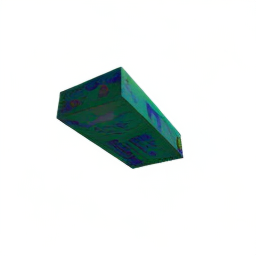}}\\
\includegraphics[width=\linewidth]{{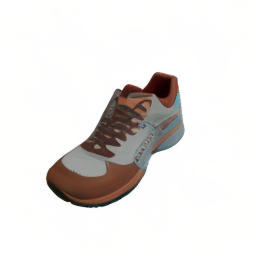}}\\
\includegraphics[width=\linewidth]{{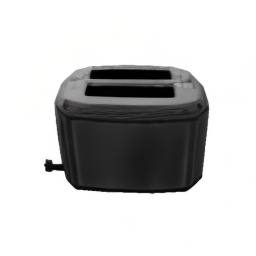}}\\
\includegraphics[width=\linewidth]{{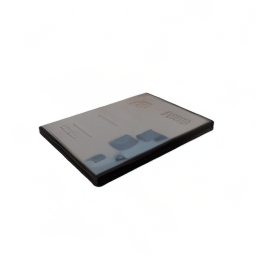}}\\
\end{minipage}
}
\caption{ Zero123 conditioning strategy.
}
\label{fig:cond}
\vspace{-5mm}
\end{figure*}

%% file: figs/supp_qvis_1.tex
\begin{figure*}[t!]
\centering
\hspace{-5mm}
\subfloat[Source View]{
\begin{minipage}[t]{0.14\linewidth}
\centering
\includegraphics[width=\linewidth]{{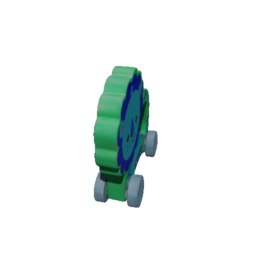}}\\

\includegraphics[width=\linewidth]{{imgs/supp/zero123_vs_valid/source/My_First_Rolling_Lion.png}}\\

\includegraphics[width=\linewidth]{{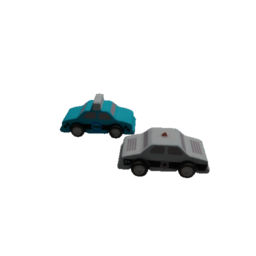}}\\

\includegraphics[width=\linewidth]{{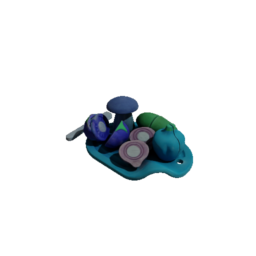}}\\

\includegraphics[width=\linewidth]{{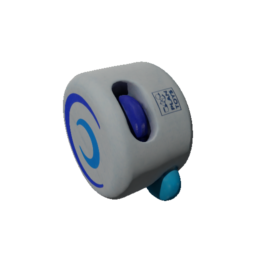}}\\

\includegraphics[width=\linewidth]{{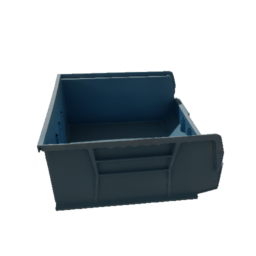}}\\

\includegraphics[width=\linewidth]{{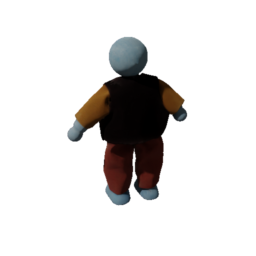}}\\

\includegraphics[width=\linewidth]{{imgs/supp/zero123_vs_valid/source/Avengers_Thor_PLlrpYniaeB.png}}\\

\end{minipage}
}
\hspace{-5mm}
\subfloat[Target View]{
\begin{minipage}[t]{0.14\linewidth}
\centering
\includegraphics[width=\linewidth]{{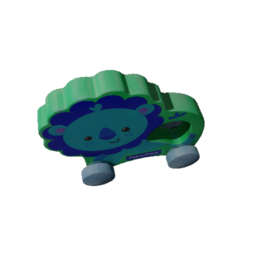}}\\

\includegraphics[width=\linewidth]{{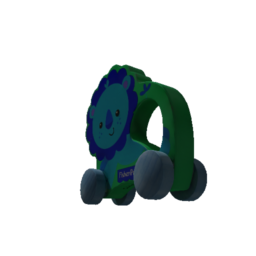}}\\

\includegraphics[width=\linewidth]{{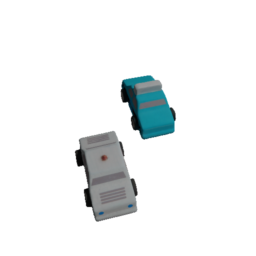}}\\

\includegraphics[width=\linewidth]{{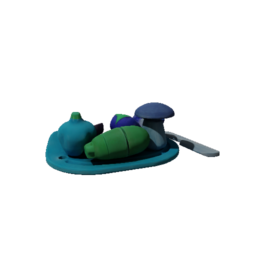}}\\

\includegraphics[width=\linewidth]{{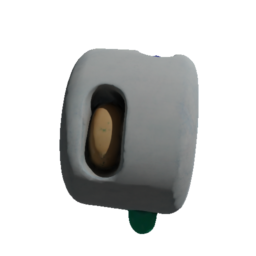}}\\

\includegraphics[width=\linewidth]{{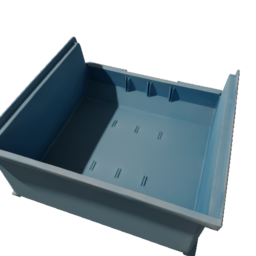}}\\

\includegraphics[width=\linewidth]{{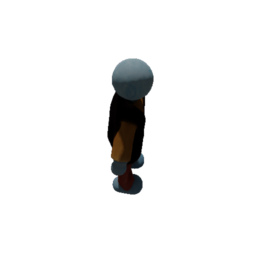}}\\

\includegraphics[width=\linewidth]{{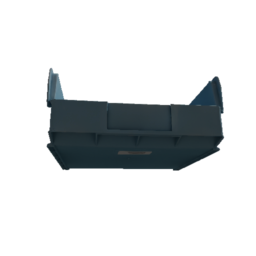}}\\

\end{minipage}
}
\hspace{-5mm}
\subfloat[Zero123\cite{liu2023zero}]{
\begin{minipage}[t]{0.14\linewidth}
\centering
\includegraphics[width=\linewidth]{{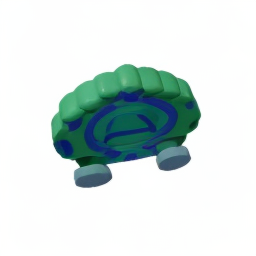}}\\

\includegraphics[width=\linewidth]{{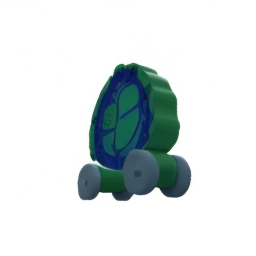}}\\

\includegraphics[width=\linewidth]{{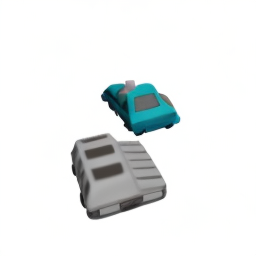}}\\

\includegraphics[width=\linewidth]{{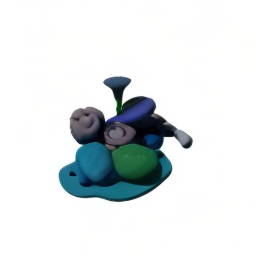}}\\

\includegraphics[width=\linewidth]{{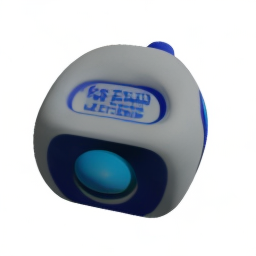}}\\

\includegraphics[width=\linewidth]{{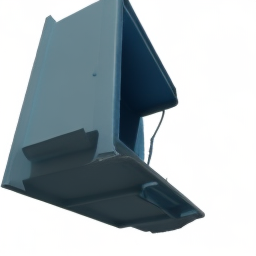}}\\

\includegraphics[width=\linewidth]{{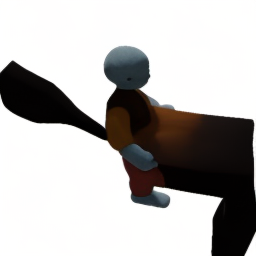}}\\

\includegraphics[width=\linewidth]{{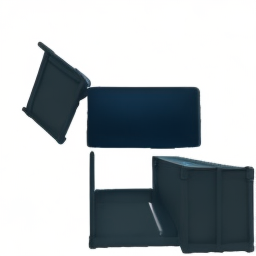}}\\
\end{minipage}
}
\hspace{-5mm}
\subfloat[VaLID (1 view)]{
\begin{minipage}[t]{0.14\linewidth}
\centering
\includegraphics[width=\linewidth]{{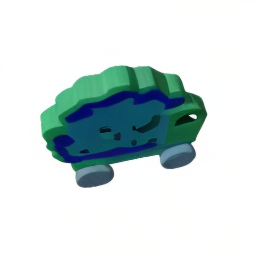}}\\

\includegraphics[width=\linewidth]{{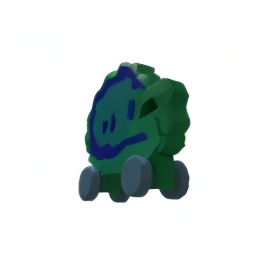}}\\
\includegraphics[width=\linewidth]{{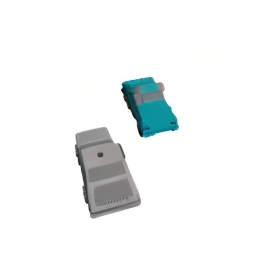}}\\

\includegraphics[width=\linewidth]{{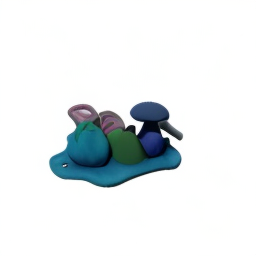}}\\

\includegraphics[width=\linewidth]{{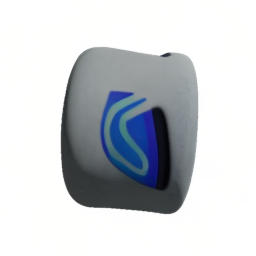}}\\

\includegraphics[width=\linewidth]{{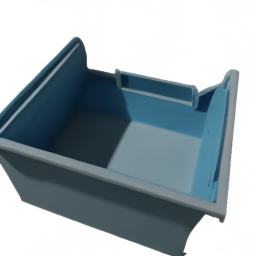}}\\

\includegraphics[width=\linewidth]{{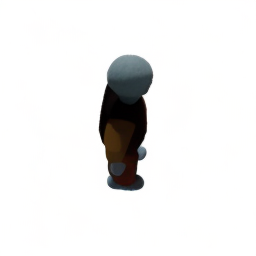}}\\

\includegraphics[width=\linewidth]{{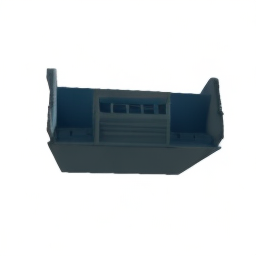}}\\

\end{minipage}
}
\hspace{-5mm}
\subfloat[VaLID (2 views)]{
\begin{minipage}[t]{0.14\linewidth}
\centering
\includegraphics[width=\linewidth]{{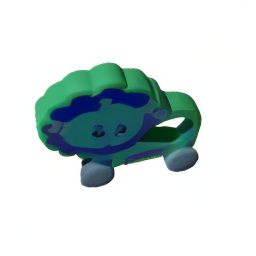}}\\

\includegraphics[width=\linewidth]{{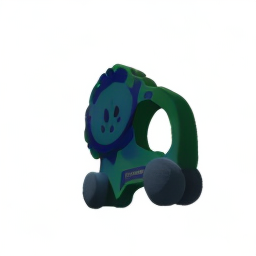}}\\
\includegraphics[width=\linewidth]{{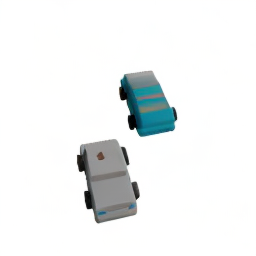}}\\

\includegraphics[width=\linewidth]{{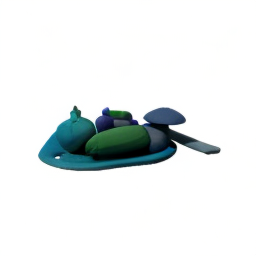}}\\

\includegraphics[width=\linewidth]{{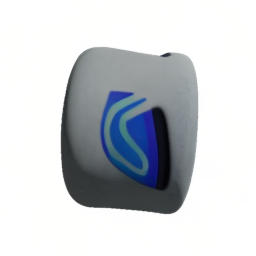}}\\

\includegraphics[width=\linewidth]{{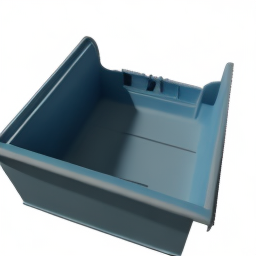}}\\

\includegraphics[width=\linewidth]{{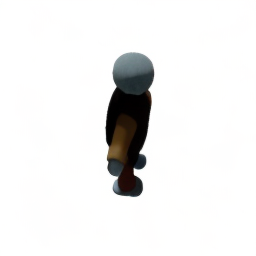}}\\

\includegraphics[width=\linewidth]{{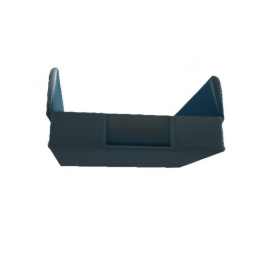}}\\

\end{minipage}
}
\hspace{-5mm}
\subfloat[VaLID (3 views)]{
\begin{minipage}[t]{0.14\linewidth}
\centering
\includegraphics[width=\linewidth]{{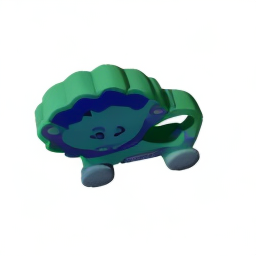}}\\

\includegraphics[width=\linewidth]{{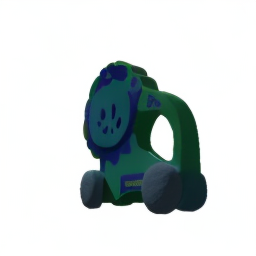}}\\

\includegraphics[width=\linewidth]{{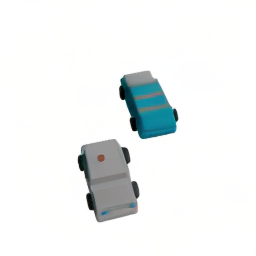}}\\

\includegraphics[width=\linewidth]{{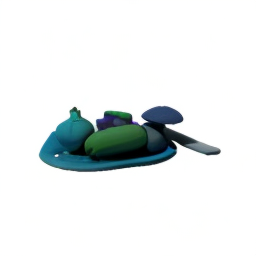}}\\

\includegraphics[width=\linewidth]{{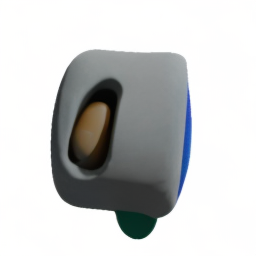}}\\

\includegraphics[width=\linewidth]{{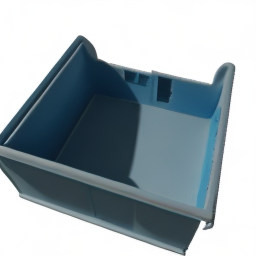}}\\

\includegraphics[width=\linewidth]{{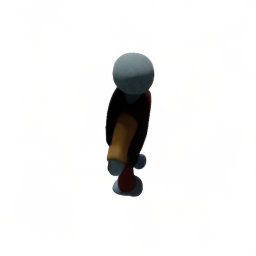}}\\

\includegraphics[width=\linewidth]{{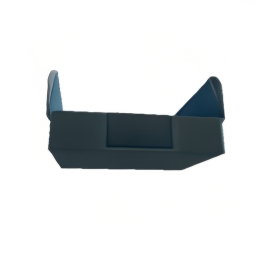}}\\

\end{minipage}
}
\hspace{-5mm}
\subfloat[VaLID (4 views)]{
\begin{minipage}[t]{0.14\linewidth}
\centering
\includegraphics[width=\linewidth]{{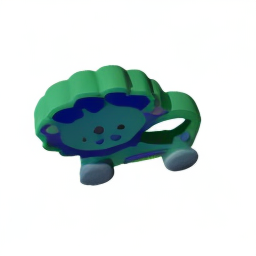}}\\

\includegraphics[width=\linewidth]{{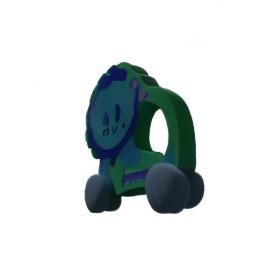}}\\

\includegraphics[width=\linewidth]{{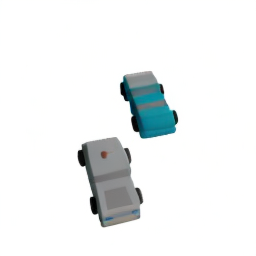}}\\

\includegraphics[width=\linewidth]{{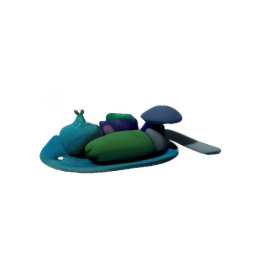}}\\

\includegraphics[width=\linewidth]{{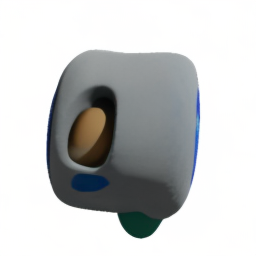}}\\

\includegraphics[width=\linewidth]{{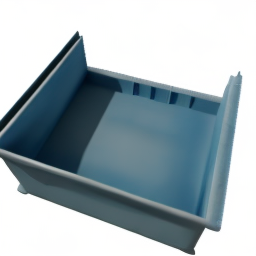}}\\

\includegraphics[width=\linewidth]{{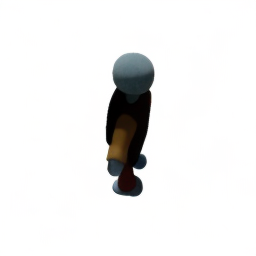}}\\

\includegraphics[width=\linewidth]{{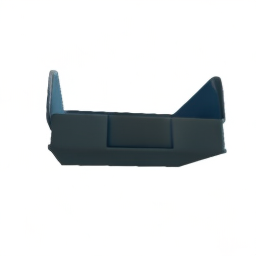}}\\

\end{minipage}
}
\caption{
More qualitative examples.
}
\label{fig:qvis_1}
\end{figure*} 

%% file: figs/supp_stage.tex
\begin{figure*}[t!]
\centering
\subfloat[Target]{
\begin{minipage}[t]{0.111\linewidth}
\centering
\includegraphics[width=\linewidth]
{{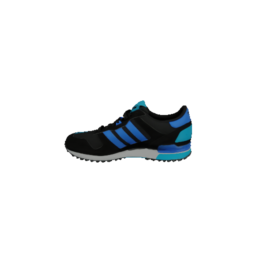}}\\
\includegraphics[width=\linewidth]
{{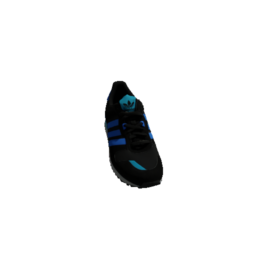}}\\
\includegraphics[width=\linewidth]
{{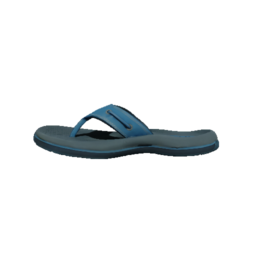}}\\
\includegraphics[width=\linewidth]
{{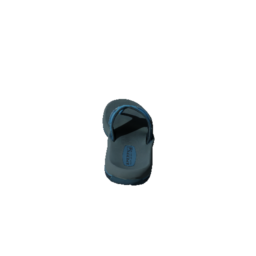}}\\
\includegraphics[width=\linewidth]
{{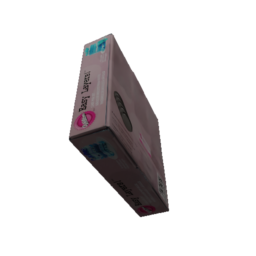}}\\
\includegraphics[width=\linewidth]
{{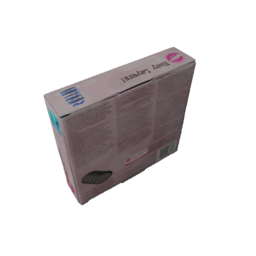}}\\
\includegraphics[width=\linewidth]
{{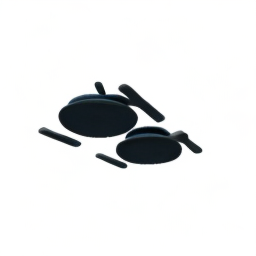}}\\
\includegraphics[width=\linewidth]
{{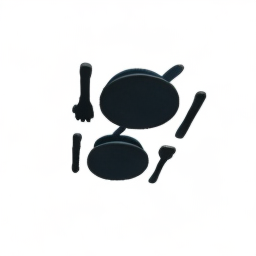}}\\
\includegraphics[width=\linewidth]
{{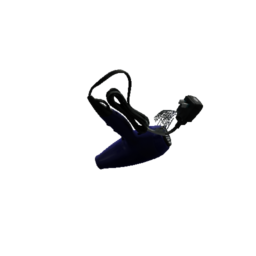}}\\
\includegraphics[width=\linewidth]
{{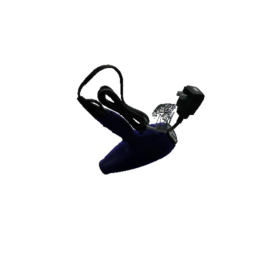}}\\
\end{minipage}
}
\subfloat[s1-1v]{
\begin{minipage}[t]{0.111\linewidth}
\centering
\includegraphics[width=\linewidth]
{{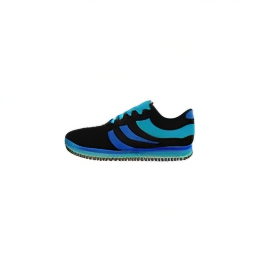}}\\
\includegraphics[width=\linewidth]
{{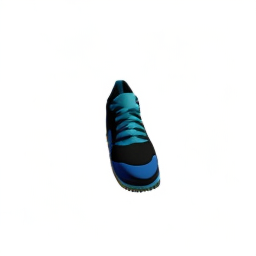}}\\
\includegraphics[width=\linewidth]
{{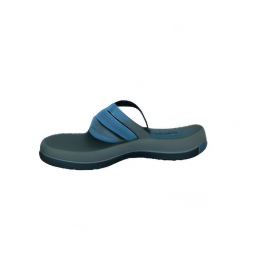}}\\
\includegraphics[width=\linewidth]
{{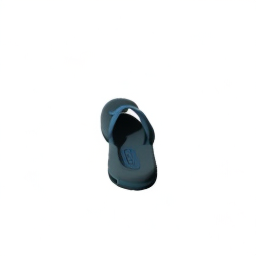}}\\
\includegraphics[width=\linewidth]
{{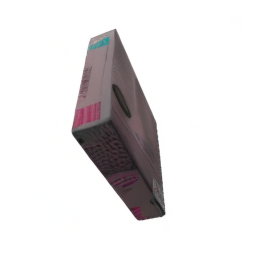}}\\
\includegraphics[width=\linewidth]
{{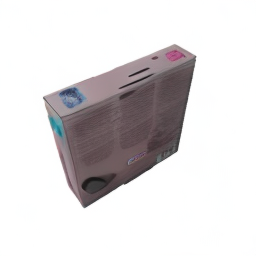}}\\
\includegraphics[width=\linewidth]
{{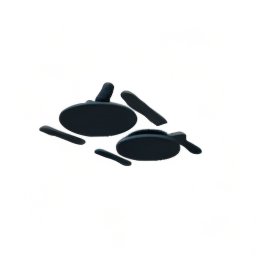}}\\
\includegraphics[width=\linewidth]
{{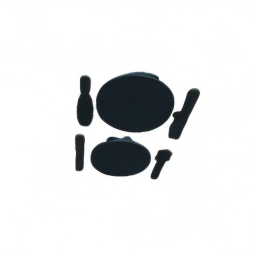}}\\
\includegraphics[width=\linewidth]
{{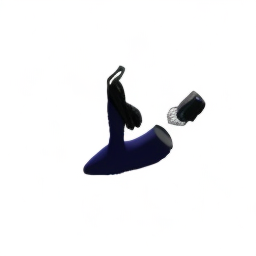}}\\
\includegraphics[width=\linewidth]
{{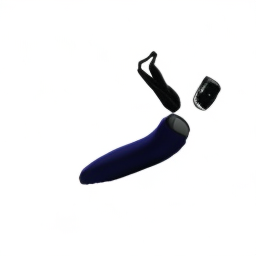}}\\
\end{minipage}
}
\hspace{-5mm}
\subfloat[s1-2v]{
\begin{minipage}[t]{0.111\linewidth}
\centering
\includegraphics[width=\linewidth]
{{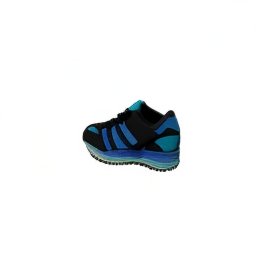}}\\
\includegraphics[width=\linewidth]
{{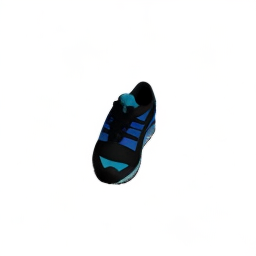}}\\
\includegraphics[width=\linewidth]
{{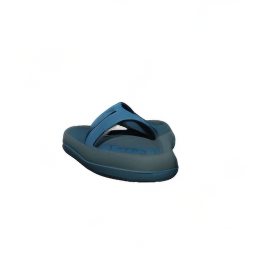}}\\
\includegraphics[width=\linewidth]
{{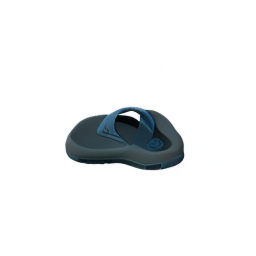}}\\
\includegraphics[width=\linewidth]
{{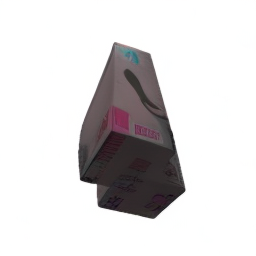}}\\
\includegraphics[width=\linewidth]
{{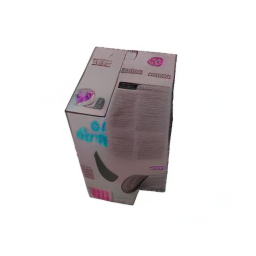}}\\
\includegraphics[width=\linewidth]
{{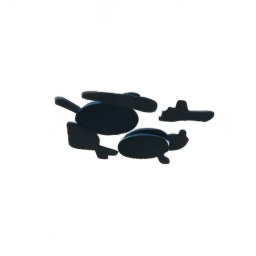}}\\
\includegraphics[width=\linewidth]
{{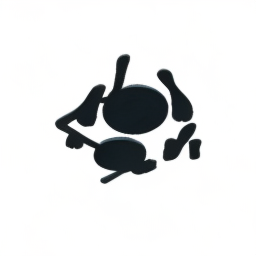}}\\
\includegraphics[width=\linewidth]
{{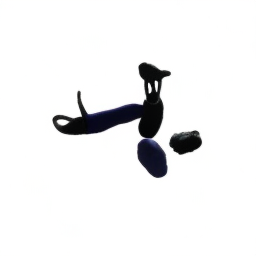}}\\
\includegraphics[width=\linewidth]
{{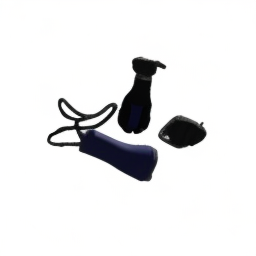}}\\
\end{minipage}
}
\hspace{-5mm}
\subfloat[s1-3v]{
\begin{minipage}[t]{0.111\linewidth}
\centering
\includegraphics[width=\linewidth]
{{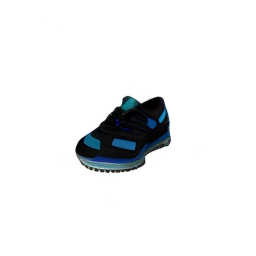}}\\
\includegraphics[width=\linewidth]
{{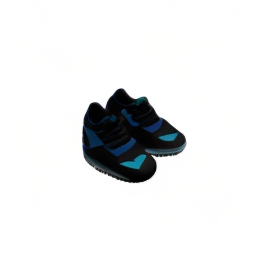}}\\
\includegraphics[width=\linewidth]
{{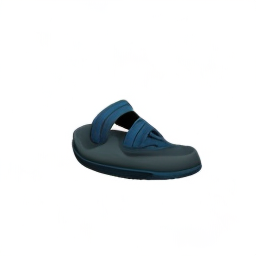}}\\
\includegraphics[width=\linewidth]
{{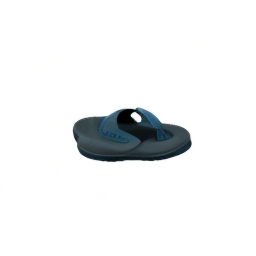}}\\
\includegraphics[width=\linewidth]
{{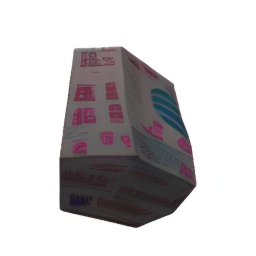}}\\
\includegraphics[width=\linewidth]
{{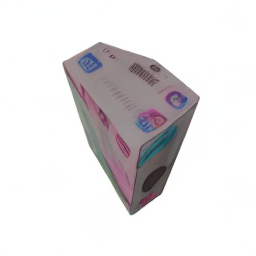}}\\
\includegraphics[width=\linewidth]
{{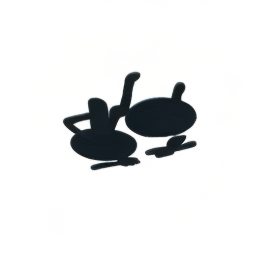}}\\
\includegraphics[width=\linewidth]
{{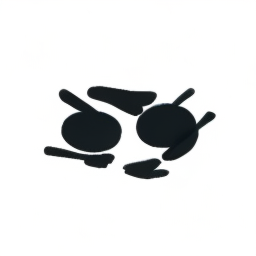}}\\
\includegraphics[width=\linewidth]
{{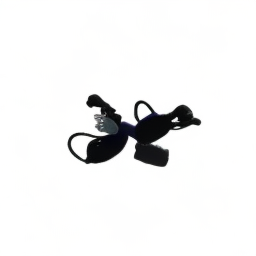}}\\
\includegraphics[width=\linewidth]
{{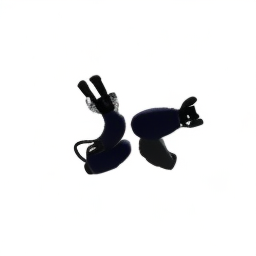}}\\
\end{minipage}
}
\hspace{-5mm}
\subfloat[s1-4v]{
\begin{minipage}[t]{0.111\linewidth}
\centering
\includegraphics[width=\linewidth]
{{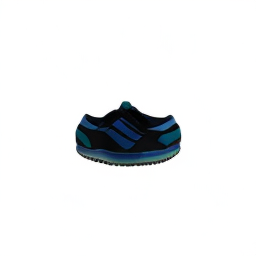}}\\
\includegraphics[width=\linewidth]
{{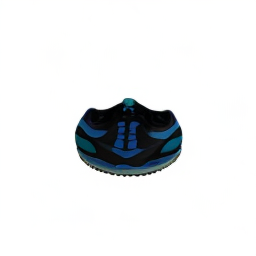}}\\
\includegraphics[width=\linewidth]
{{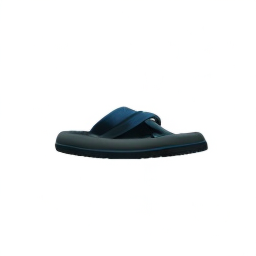}}\\
\includegraphics[width=\linewidth]
{{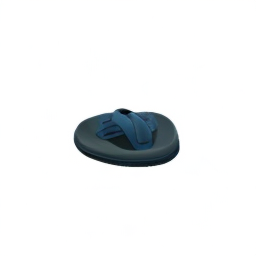}}\\
\includegraphics[width=\linewidth]
{{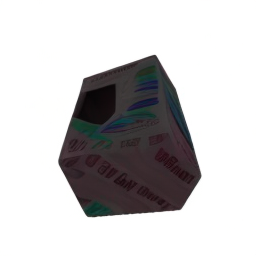}}\\
\includegraphics[width=\linewidth]
{{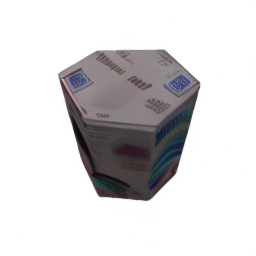}}\\
\includegraphics[width=\linewidth]
{{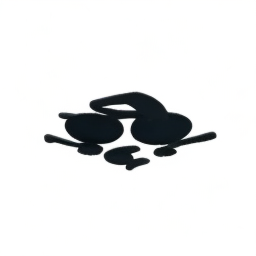}}\\
\includegraphics[width=\linewidth]
{{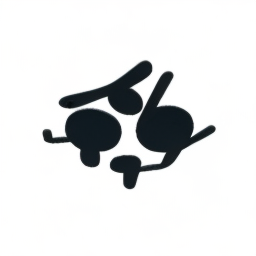}}\\
\includegraphics[width=\linewidth]
{{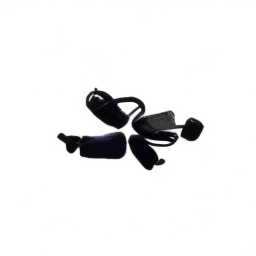}}\\
\includegraphics[width=\linewidth]
{{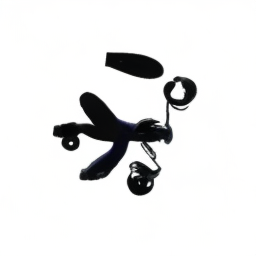}}\\
\end{minipage}
}
\subfloat[s2-1v]{
\begin{minipage}[t]{0.111\linewidth}
\centering
\includegraphics[width=\linewidth]
{{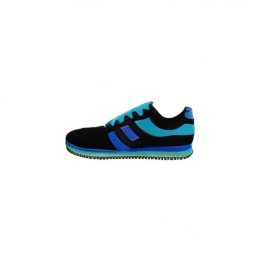}}\\
\includegraphics[width=\linewidth]
{{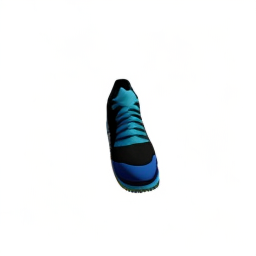}}\\
\includegraphics[width=\linewidth]
{{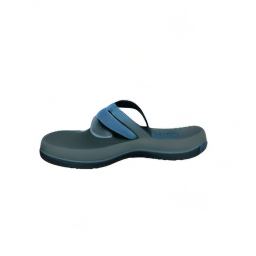}}\\
\includegraphics[width=\linewidth]
{{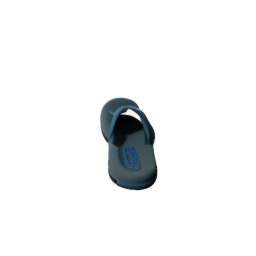}}\\
\includegraphics[width=\linewidth]
{{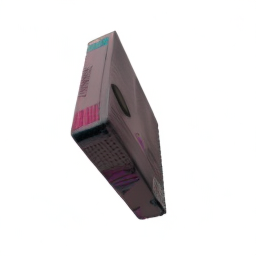}}\\
\includegraphics[width=\linewidth]
{{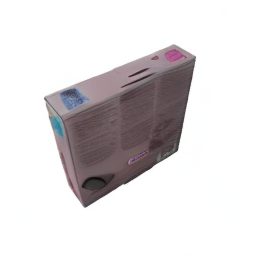}}\\
\includegraphics[width=\linewidth]
{{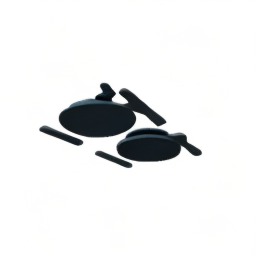}}\\
\includegraphics[width=\linewidth]
{{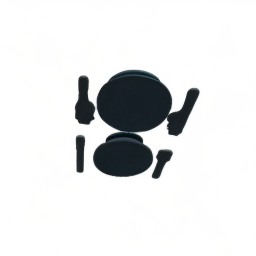}}\\
\includegraphics[width=\linewidth]
{{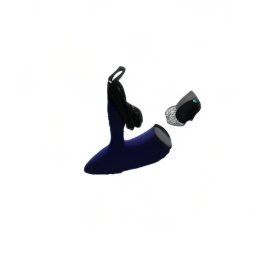}}\\
\includegraphics[width=\linewidth]
{{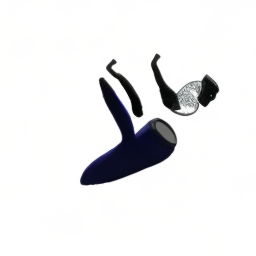}}\\
\end{minipage}
}
\hspace{-5mm}
\subfloat[s2-2v]{
\begin{minipage}[t]{0.111\linewidth}
\centering
\includegraphics[width=\linewidth]
{{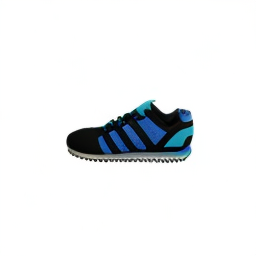}}\\
\includegraphics[width=\linewidth]
{{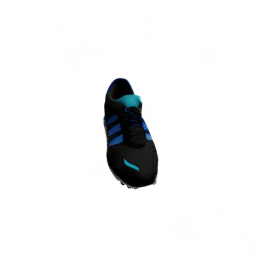}}\\
\includegraphics[width=\linewidth]
{{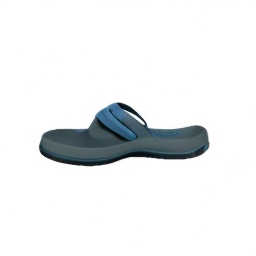}}\\
\includegraphics[width=\linewidth]
{{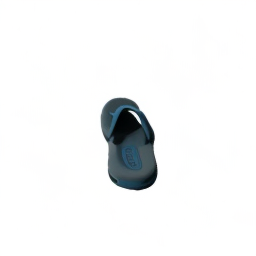}}\\
\includegraphics[width=\linewidth]
{{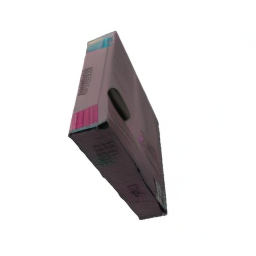}}\\
\includegraphics[width=\linewidth]
{{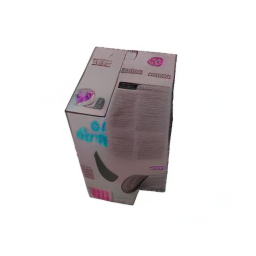}}\\
\includegraphics[width=\linewidth]
{{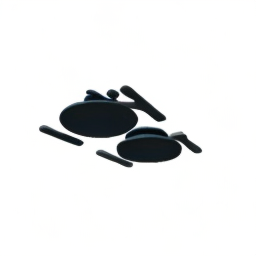}}\\
\includegraphics[width=\linewidth]
{{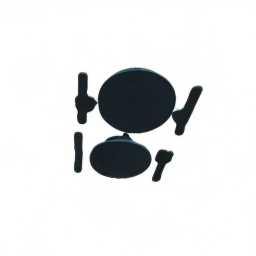}}\\
\includegraphics[width=\linewidth]
{{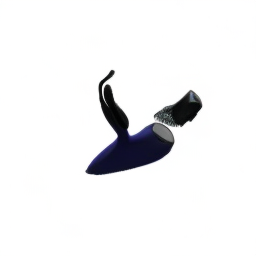}}\\
\includegraphics[width=\linewidth]
{{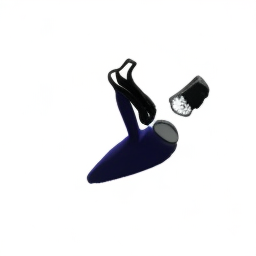}}\\
\end{minipage}
}
\hspace{-5mm}
\subfloat[s2-3v]{
\begin{minipage}[t]{0.111\linewidth}
\centering
\includegraphics[width=\linewidth]
{{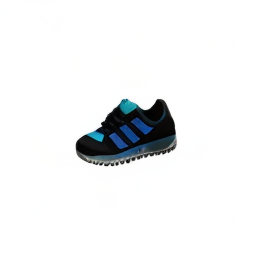}}\\
\includegraphics[width=\linewidth]
{{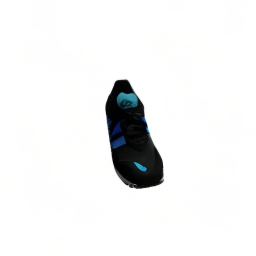}}\\
\includegraphics[width=\linewidth]
{{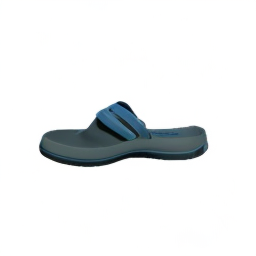}}\\
\includegraphics[width=\linewidth]
{{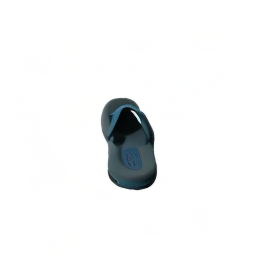}}\\
\includegraphics[width=\linewidth]
{{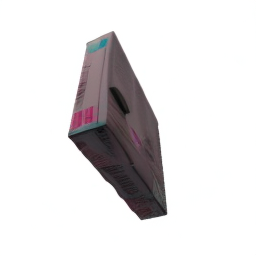}}\\
\includegraphics[width=\linewidth]
{{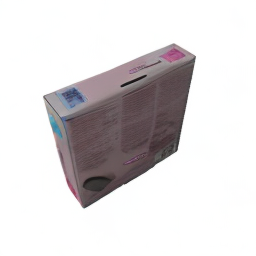}}\\
\includegraphics[width=\linewidth]
{{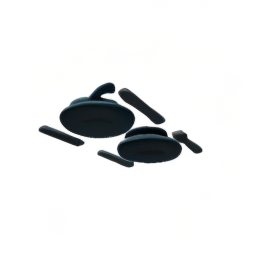}}\\
\includegraphics[width=\linewidth]
{{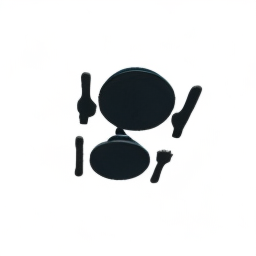}}\\
\includegraphics[width=\linewidth]
{{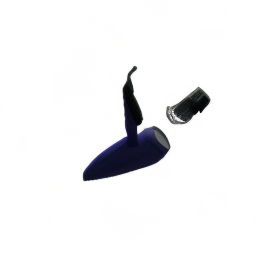}}\\
\includegraphics[width=\linewidth]
{{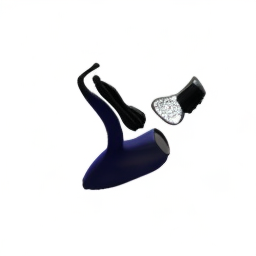}}\\
\end{minipage}
}
\hspace{-5mm}
\subfloat[s2-4v]{
\begin{minipage}[t]{0.111\linewidth}
\centering
\includegraphics[width=\linewidth]
{{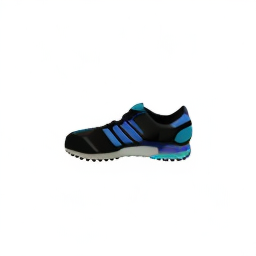}}\\
\includegraphics[width=\linewidth]
{{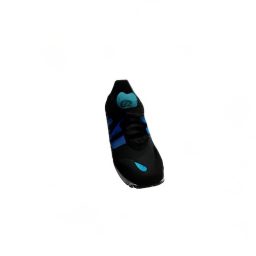}}\\
\includegraphics[width=\linewidth]
{{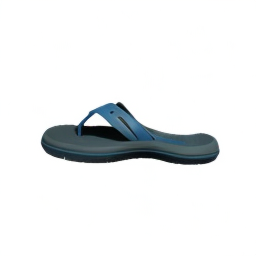}}\\
\includegraphics[width=\linewidth]
{{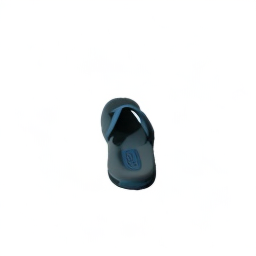}}\\
\includegraphics[width=\linewidth]
{{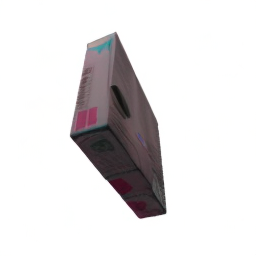}}\\
\includegraphics[width=\linewidth]
{{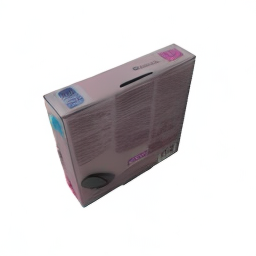}}\\
\includegraphics[width=\linewidth]
{{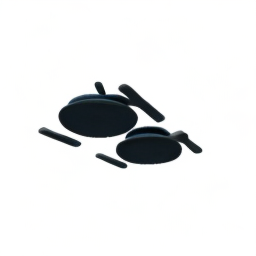}}\\
\includegraphics[width=\linewidth]
{{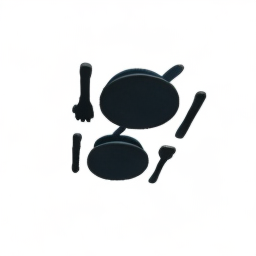}}\\
\includegraphics[width=\linewidth]
{{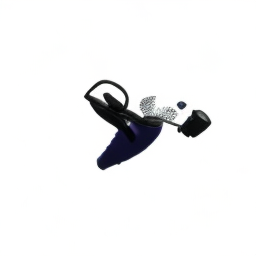}}\\
\includegraphics[width=\linewidth]
{{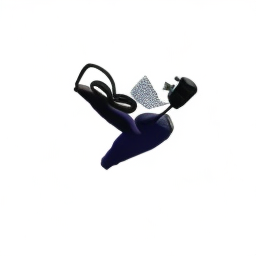}}\\
\end{minipage}
}
\caption{
Impact of stage 2 training of the proposed VaLID method. Columns (b)-(e) show the inference results on the variable number of input views (up to 4 views) after stage 1 training (s1). Columns (f)-(i) show the inference results after stage 2 training (s2) when the Cross Former parameters are tuned to perform multi-view image fusion.}
\label{fig:stage}
\vspace{-5mm}
\end{figure*}

%% file: cvpr_arxiv.bbl
\begin{thebibliography}{10}\itemsep=-1pt

\bibitem{sdiv}
Stable diffusion image variations - a hugging face space by lambdalabs.

\bibitem{Chan2022}
Eric~R. Chan, Connor~Z. Lin, Matthew~A. Chan, Koki Nagano, Boxiao Pan,
  Shalini~De Mello, Orazio Gallo, Leonidas Guibas, Jonathan Tremblay, Sameh
  Khamis, Tero Karras, and Gordon Wetzstein.
\newblock Efficient geometry-aware {3D} generative adversarial networks.
\newblock In {\em CVPR}, 2022.

\bibitem{deitke2023objaverse}
Matt Deitke, Dustin Schwenk, Jordi Salvador, Luca Weihs, Oscar Michel, Eli
  VanderBilt, Ludwig Schmidt, Kiana Ehsani, Aniruddha Kembhavi, and Ali
  Farhadi.
\newblock Objaverse: A universe of annotated 3d objects.
\newblock In {\em Proceedings of the IEEE conference on Computer Vision and
  Pattern Recognition}, pages 13142--13153, 2023.

\bibitem{dosovitskiy2020image}
Alexey Dosovitskiy, Lucas Beyer, Alexander Kolesnikov, Dirk Weissenborn,
  Xiaohua Zhai, Thomas Unterthiner, Mostafa Dehghani, Matthias Minderer, Georg
  Heigold, Sylvain Gelly, et~al.
\newblock An image is worth 16x16 words: Transformers for image recognition at
  scale.
\newblock {\em arXiv preprint arXiv:2010.11929}, 2020.

\bibitem{downs2022google}
Laura Downs, Anthony Francis, Nate Koenig, Brandon Kinman, Ryan Hickman, Krista
  Reymann, Thomas~B McHugh, and Vincent Vanhoucke.
\newblock Google scanned objects: A high-quality dataset of 3d scanned
  household items.
\newblock In {\em International Conference on Robotics and Automation}, pages
  2553--2560. IEEE, 2022.

\bibitem{dupont2020equivariant}
Emilien Dupont, Bautista Miguel~Angel, Alex Colburn, Aditya Sankar, Carlos
  Guestrin, Josh Susskind, and Qi Shan.
\newblock Equivariant neural rendering.
\newblock In {\em International Conference on Machine Learning}, 2020.

\bibitem{goodfellow2014generative}
Ian Goodfellow, Jean Pouget-Abadie, Mehdi Mirza, Bing Xu, David Warde-Farley,
  Sherjil Ozair, Aaron Courville, and Yoshua Bengio.
\newblock Generative adversarial nets.
\newblock In {\em Advances in neural information processing systems}, pages
  2672--2680, 2014.

\bibitem{han2022single}
Yuxuan Han, Ruicheng Wang, and Jiaolong Yang.
\newblock Single-view view synthesis in the wild with learned adaptive
  multiplane images.
\newblock In {\em ACM SIGGRAPH}, 2022.

\bibitem{he2022masked}
Kaiming He, Xinlei Chen, Saining Xie, Yanghao Li, Piotr Doll{\'a}r, and Ross
  Girshick.
\newblock Masked autoencoders are scalable vision learners.
\newblock In {\em Proceedings of the IEEE/CVF conference on computer vision and
  pattern recognition}, pages 16000--16009, 2022.

\bibitem{henzler2021UnsupervisedLO}
Philipp Henzler, Jeremy Reizenstein, Patrick Labatut, Roman Shapovalov, Tobias
  Ritschel, Andrea Vedaldi, and David Novotn{\'y}.
\newblock Unsupervised learning of 3d object categories from videos in the
  wild.
\newblock {\em 2021 IEEE/CVF Conference on Computer Vision and Pattern
  Recognition (CVPR)}, pages 4698--4707, 2021.

\bibitem{ho2020denoising}
Jonathan Ho, Ajay Jain, and Pieter Abbeel.
\newblock Denoising diffusion probabilistic models.
\newblock {\em Advances in neural information processing systems},
  33:6840--6851, 2020.

\bibitem{jain2021putting}
Ajay Jain, Matthew Tancik, and Pieter Abbeel.
\newblock Putting nerf on a diet: Semantically consistent few-shot view
  synthesis.
\newblock In {\em IEEE International Conference on Computer Vision}, pages
  5885--5894, 2021.

\bibitem{kant2023invs}
Yash Kant, Aliaksandr Siarohin, Michael Vasilkovsky, Riza~Alp Guler, Jian Ren,
  Sergey Tulyakov, and Igor Gilitschenski.
\newblock invs : Repurposing diffusion inpainters for novel view synthesis.
\newblock In {\em SIGGRAPH Asia 2023 Conference Papers}, 2023.

\bibitem{kopf2020one}
Johannes Kopf, Kevin Matzen, Suhib Alsisan, Ocean Quigley, Francis Ge, Yangming
  Chong, Josh Patterson, Jan-Michael Frahm, Shu Wu, Matthew Yu, et~al.
\newblock One shot 3d photography.
\newblock {\em ACM Transactions on Graphics (TOG)}, 39(4):76--1, 2020.

\bibitem{lin2023magic3d}
Chen-Hsuan Lin, Jun Gao, Luming Tang, Towaki Takikawa, Xiaohui Zeng, Xun Huang,
  Karsten Kreis, Sanja Fidler, Ming-Yu Liu, and Tsung-Yi Lin.
\newblock Magic3d: High-resolution text-to-3d content creation.
\newblock In {\em IEEE Conference on Computer Vision and Pattern Recognition
  ({CVPR})}, 2023.

\bibitem{lin2023visionnerf}
Kai-En Lin, Lin Yen-Chen, Wei-Sheng Lai, Tsung-Yi Lin, Yi-Chang Shih, and Ravi
  Ramamoorthi.
\newblock Vision transformer for nerf-based view synthesis from a single input
  image.
\newblock In {\em WACV}, 2023.

\bibitem{lin2023consistent123}
Yukang Lin, Haonan Han, Chaoqun Gong, Zunnan Xu, Yachao Zhang, and Xiu Li.
\newblock Consistent123: One image to highly consistent 3d asset using
  case-aware diffusion priors.
\newblock {\em arXiv preprint arXiv:2309.17261}, 2023.

\bibitem{liu2023zero}
Ruoshi Liu, Rundi Wu, Basile Van~Hoorick, Pavel Tokmakov, Sergey Zakharov, and
  Carl Vondrick.
\newblock Zero-1-to-3: Zero-shot one image to 3d object.
\newblock {\em arXiv preprint arXiv:2303.11328}, 2023.

\bibitem{liu2023syncdreamer}
Yuan Liu, Cheng Lin, Zijiao Zeng, Xiaoxiao Long, Lingjie Liu, Taku Komura, and
  Wenping Wang.
\newblock Syncdreamer: Learning to generate multiview-consistent images from a
  single-view image.
\newblock {\em arXiv preprint arXiv:2309.03453}, 2023.

\bibitem{long2023wonder3d}
Xiaoxiao Long, Yuan-Chen Guo, Cheng Lin, Yuan Liu, Zhiyang Dou, Lingjie Liu,
  Yuexin Ma, Song-Hai Zhang, Marc Habermann, Christian Theobalt, and Wenping
  Wang.
\newblock Wonder3d: Single image to 3d using cross-domain diffusion.
\newblock {\em arXiv:2310.15008}, 2023.

\bibitem{melaskyriazi2023realfusion}
Luke Melas-Kyriazi, Christian Rupprecht, Iro Laina, and Andrea Vedaldi.
\newblock Realfusion: 360 reconstruction of any object from a single image.
\newblock In {\em CVPR}, 2023.

\bibitem{mildenhall2021nerf}
Ben Mildenhall, Pratul~P Srinivasan, Matthew Tancik, Jonathan~T Barron, Ravi
  Ramamoorthi, and Ren Ng.
\newblock Nerf: Representing scenes as neural radiance fields for view
  synthesis.
\newblock {\em Communications of the ACM}, 65(1):99--106, 2021.

\bibitem{mu20223d}
Fangzhou Mu, Jian Wang, Yicheng Wu, and Yin Li.
\newblock 3d photo stylization: Learning to generate stylized novel views from
  a single image.
\newblock In {\em Proceedings of the IEEE/CVF Conference on Computer Vision and
  Pattern Recognition}, pages 16273--16282, 2022.

\bibitem{niemeyer2022regnerf}
Michael Niemeyer, Jonathan~T Barron, Ben Mildenhall, Mehdi~SM Sajjadi, Andreas
  Geiger, and Noha Radwan.
\newblock Regnerf: Regularizing neural radiance fields for view synthesis from
  sparse inputs.
\newblock In {\em Proceedings of the IEEE/CVF Conference on Computer Vision and
  Pattern Recognition}, pages 5480--5490, 2022.

\bibitem{poole2022dreamfusion}
Ben Poole, Ajay Jain, Jonathan~T Barron, and Ben Mildenhall.
\newblock Dreamfusion: Text-to-3d using 2d diffusion.
\newblock In {\em International Conference on Learning Representations}, 2023.

\bibitem{radford2021learning}
Alec Radford, Jong~Wook Kim, Chris Hallacy, Aditya Ramesh, Gabriel Goh,
  Sandhini Agarwal, Girish Sastry, Amanda Askell, Pamela Mishkin, Jack Clark,
  et~al.
\newblock Learning transferable visual models from natural language
  supervision.
\newblock In {\em International conference on machine learning}, pages
  8748--8763. PMLR, 2021.

\bibitem{rombach2022high}
Robin Rombach, Andreas Blattmann, Dominik Lorenz, Patrick Esser, and Bj{\"o}rn
  Ommer.
\newblock High-resolution image synthesis with latent diffusion models.
\newblock In {\em Proceedings of the IEEE/CVF conference on computer vision and
  pattern recognition}, pages 10684--10695, 2022.

\bibitem{sajjadi2022srt}
Mehdi S.~M. Sajjadi, Henning Meyer, Etienne Pot, Urs Bergmann, Klaus Greff,
  Noha Radwan, Suhani Vora, Mario Lucic, Daniel Duckworth, Alexey Dosovitskiy,
  Jakob Uszkoreit, Thomas Funkhouser, and Andrea Tagliasacchi.
\newblock {Scene Representation Transformer: Geometry-Free Novel View Synthesis
  Through Set-Latent Scene Representations}.
\newblock In {\em Proceedings of the IEEE conference on Computer Vision and
  Pattern Recognition}, 2022.

\bibitem{shi2023MVDream}
Yichun Shi, Peng Wang, Jianglong Ye, Long Mai, Kejie Li, and Xiao Yang.
\newblock Mvdream: Multi-view diffusion for 3d generation.
\newblock {\em arXiv:2308.16512}, 2023.

\bibitem{sitzmann2021lfns}
Vincent Sitzmann, Semon Rezchikov, William~T. Freeman, Joshua~B. Tenenbaum, and
  Fredo Durand.
\newblock Light field networks: Neural scene representations with
  single-evaluation rendering.
\newblock In {\em Neural Information Processing Systems}, 2021.

\bibitem{sohl2015diffusion}
Jascha Sohl-Dickstein, Eric~A. Weiss, Niru Maheswaranathan, and Surya Ganguli.
\newblock {Deep Unsupervised Learning using Nonequilibrium Thermodynamics}.
\newblock In {\em {Proceedings of the 32nd International Conference on Machine
  Learning}}, {JMLR Proceedings}, pages 2256--2265, 2015.

\bibitem{szymanowicz23viewset_diffusion}
Stanislaw Szymanowicz, Christian Rupprecht, and Andrea Vedaldi.
\newblock Viewset diffusion: (0-)image-conditioned 3d generative models from 2d
  data.
\newblock {\em International Conference on Computer Vision}, 2023.

\bibitem{tang2023mvdiffusion}
Shitao Tang, Fuyang Zhang, Jiacheng Chen, Peng Wang, and Yasutaka Furukawa.
\newblock Mvdiffusion: Enabling holistic multi-view image generation with
  correspondence-aware diffusion.
\newblock In {\em Neural Information Processing Systems}, 2023.

\bibitem{tremblay2022rtmv}
Jonathan Tremblay, Moustafa Meshry, Alex Evans, Jan Kautz, Alexander Keller,
  Sameh Khamis, Thomas M{\"u}ller, Charles Loop, Nathan Morrical, Koki Nagano,
  et~al.
\newblock Rtmv: A ray-traced multi-view synthetic dataset for novel view
  synthesis.
\newblock {\em arXiv preprint arXiv:2205.07058}, 2022.

\bibitem{tseng2023poseguideddiffusion}
Hung-Yu Tseng, Qinbo Li, Changil Kim, Suhib Alsisan, Jia-Bin Huang, and
  Johannes Kopf.
\newblock Consistent view synthesis with pose-guided diffusion models.
\newblock In {\em Proceedings of the IEEE conference on Computer Vision and
  Pattern Recognition}, 2023.

\bibitem{tucker2020singleviewVS}
Richard Tucker and Noah Snavely.
\newblock Single-view view synthesis with multiplane images.
\newblock {\em 2020 IEEE/CVF Conference on Computer Vision and Pattern
  Recognition (CVPR)}, pages 548--557, 2020.

\bibitem{wang2023score}
Haochen Wang, Xiaodan Du, Jiahao Li, Raymond~A Yeh, and Greg Shakhnarovich.
\newblock Score jacobian chaining: Lifting pretrained 2d diffusion models for
  3d generation.
\newblock In {\em Proceedings of the IEEE conference on Computer Vision and
  Pattern Recognition}, pages 12619--12629, 2023.

\bibitem{wang2021ibrnet}
Qianqian Wang, Zhicheng Wang, Kyle Genova, Pratul Srinivasan, Howard Zhou,
  Jonathan~T. Barron, Ricardo Martin-Brualla, Noah Snavely, and Thomas
  Funkhouser.
\newblock Ibrnet: Learning multi-view image-based rendering.
\newblock In {\em CVPR}, 2021.

\bibitem{wang2004image}
Zhou Wang, Alan~C Bovik, Hamid~R Sheikh, and Eero~P Simoncelli.
\newblock Image quality assessment: from error visibility to structural
  similarity.
\newblock {\em IEEE Transactions on Image Processing}, 13(4):600--612, 2004.

\bibitem{watson2022novel}
Daniel Watson, William Chan, Ricardo Martin-Brualla, Jonathan Ho, Andrea
  Tagliasacchi, and Mohammad Norouzi.
\newblock Novel view synthesis with diffusion models.
\newblock In {\em International Conference on Learning Representations}, 2023.

\bibitem{weng2023consistent123}
Haohan Weng, Tianyu Yang, Jianan Wang, Yu Li, Tong Zhang, C.~L.~Philip Chen,
  and Lei Zhang.
\newblock Consistent123: Improve consistency for one image to 3d object
  synthesis.
\newblock {\em arXiv preprint arXiv:2310.08092}, 2023.

\bibitem{xuneurallift360}
Dejia Xu, Yifan~Jiang 0001, Peihao Wang, Zhiwen Fan, Yi Wang, and Zhangyang
  Wang.
\newblock Neurallift-360: Lifting an in-the-wild 2d photo to a 3d object with
  360° views.
\newblock In {\em IEEE/CVF Conference on Computer Vision and Pattern
  Recognition, CVPR 2023, Vancouver, BC, Canada, June 17-24, 2023}, pages
  4479--4489. IEEE, 2023.

\bibitem{xu2022sinnerf}
Dejia Xu, Yifan Jiang, Peihao Wang, Zhiwen Fan, Humphrey Shi, and Zhangyang
  Wang.
\newblock Sinnerf: Training neural radiance fields on complex scenes from a
  single image.
\newblock In {\em Computer Vision--ECCV 2022: 17th European Conference, Tel
  Aviv, Israel, October 23--27, 2022, Proceedings, Part XXII}, pages 736--753.
  Springer, 2022.

\bibitem{yang2023consistnet}
Jiayu Yang, Ziang Cheng, Yunfei Duan, Pan Ji, and Hongdong Li.
\newblock Consistnet: Enforcing 3d consistency for multi-view images diffusion.
\newblock {\em arXiv}, 2023.

\bibitem{yariv2021volume}
Lior Yariv, Jiatao Gu, Yoni Kasten, and Yaron Lipman.
\newblock Volume rendering of neural implicit surfaces.
\newblock In {\em Thirty-Fifth Conference on Neural Information Processing
  Systems}, 2021.

\bibitem{ye2023consistent}
Jianglong Ye, Peng Wang, Kejie Li, Yichun Shi, and Heng Wang.
\newblock Consistent-1-to-3: Consistent image to 3d view synthesis via
  geometry-aware diffusion models.
\newblock {\em arXiv preprint arXiv:2310.03020}, 2023.

\bibitem{yen2022nerf}
Lin Yen-Chen, Pete Florence, Jonathan~T Barron, Tsung-Yi Lin, Alberto
  Rodriguez, and Phillip Isola.
\newblock Nerf-supervision: Learning dense object descriptors from neural
  radiance fields.
\newblock In {\em 2022 International Conference on Robotics and Automation
  (ICRA)}, pages 6496--6503. IEEE, 2022.

\bibitem{yu2021pixelnerf}
Alex Yu, Vickie Ye, Matthew Tancik, and Angjoo Kanazawa.
\newblock {pixelNeRF}: Neural radiance fields from one or few images.
\newblock In {\em CVPR}, 2021.

\bibitem{zhang2018unreasonable}
Richard Zhang, Phillip Isola, Alexei~A Efros, Eli Shechtman, and Oliver Wang.
\newblock The unreasonable effectiveness of deep features as a perceptual
  metric.
\newblock In {\em Proceedings of the IEEE conference on Computer Vision and
  Pattern Recognition}, pages 586--595, 2018.

\end{thebibliography}
